\pdfoutput=1
\documentclass[10pt,twocolumn,letterpaper]{article}

\usepackage[pagenumbers]{cvpr}

\usepackage[dvipsnames]{xcolor}
\usepackage{amsmath}
\usepackage{amssymb}
\usepackage{graphicx}
\usepackage{booktabs}
\usepackage{mathtools}
\usepackage[normalem]{ulem}
\definecolor{cvprblue}{rgb}{0.21,0.49,0.74}
\usepackage[pagebackref,breaklinks,colorlinks,allcolors=cvprblue]{hyperref}

\title{Generalizing Agentic Robotics via Multi-Teacher On-Policy Distillation}

\author{%
Xinxin Liu\textsuperscript{1}\quad Anh Dao\textsuperscript{1}\quad
Wenrui Bao\textsuperscript{1}\quad Rafid Ahmed\textsuperscript{1}\\
Bingxin Xu\textsuperscript{2}\quad Ming Li\textsuperscript{1}\quad
Chen Chen\textsuperscript{1}\quad Yuzhang Shang\textsuperscript{1}\\[0.3em]
\textsuperscript{1}University of Central Florida\quad
\textsuperscript{2}University of Southern California
}

\begin{document}
\maketitle
\begin{abstract}
Building general-purpose robots is a central goal in robotics. Agentic robotics pursues this goal with planners that control robots through tool calls. However, existing methods either rely on proprietary models accessed through APIs, which incur high costs and latency, or post-train a planner with reinforcement learning (RL) in a single domain, requiring a separate specialist for each domain with incompatible tools. In large language model (LLM) post-training, multi-teacher on-policy distillation (MOPD) has been used to consolidate domain specialists into a single generalist. Yet whether MOPD can build a generalist planner for agentic robotics, how well it performs, and how it compares with other consolidation methods remain unexplored. In this paper, we systematically study these questions: we introduce \textbf{Robo-MOPD}, which trains a single student on closed-loop rollouts in simulation under token-level supervision from matching frozen specialists, and compare it with weight merging, off-policy specialist supervised fine-tuning (SFT), and multi-domain online RL. Experiments on LIBERO-PRO and RoboCasa365 show that a single 9B Robo-MOPD planner reaches 72.9\% and 44.8\% success (58.9\% average), surpassing all consolidated baselines in average success and both 9B specialists on their respective benchmarks. Robo-MOPD excels on long-horizon tasks, outperforming multi-domain online RL by 13.0 percentage points on \mbox{LIBERO-10}.
\end{abstract}

\section{Introduction}
\label{sec:intro}

\begin{figure}[!t]
  \centering
  \includegraphics[width=\columnwidth]{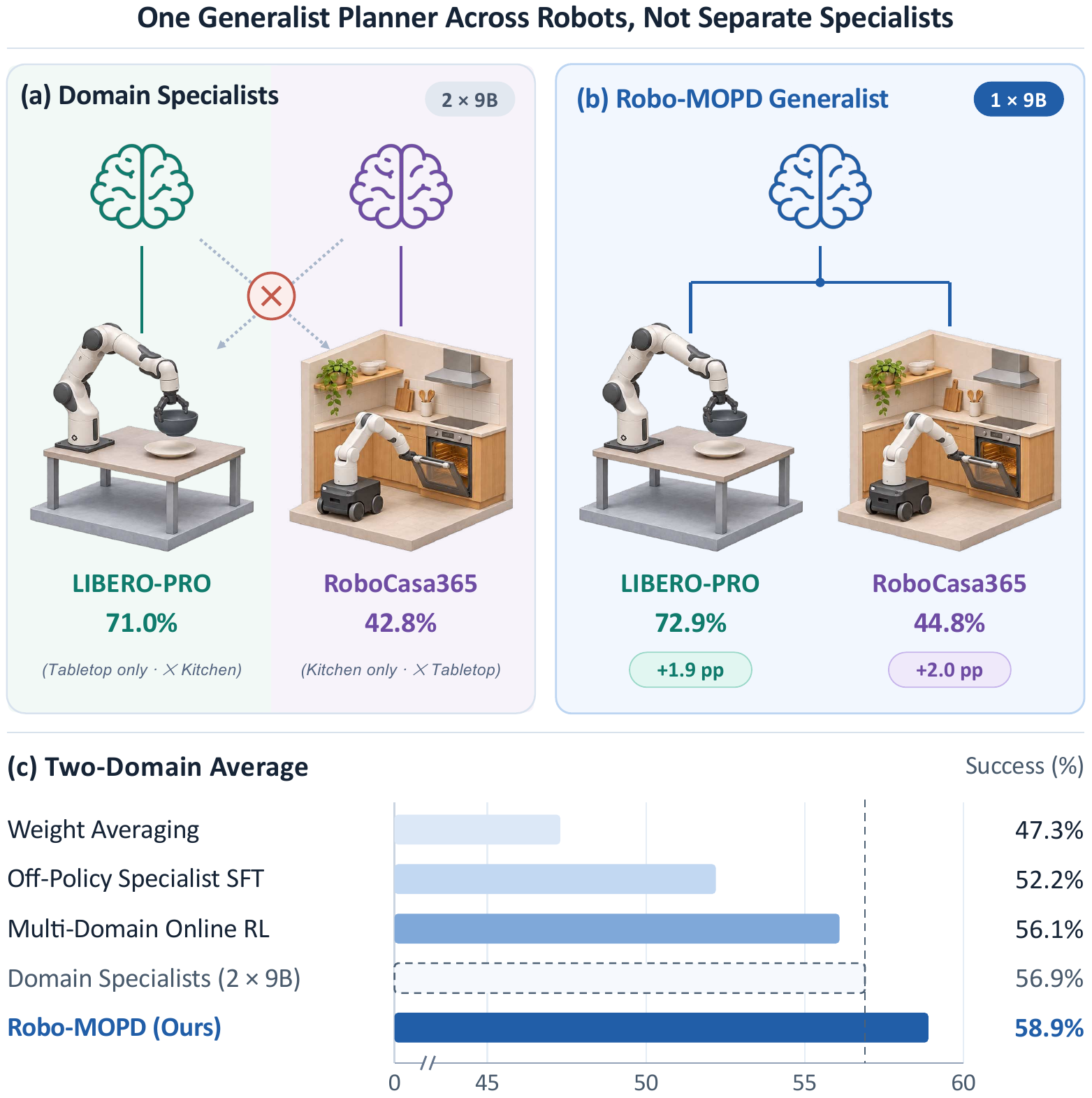}
  \vspace{-6mm}
  \caption{\textbf{One generalist planner vs.\ separate domain specialists.} (a) Domain specialists require a separate 9B model per domain and cannot operate across incompatible tools. (b) \textbf{Robo-MOPD} consolidates them into one 9B generalist spanning LIBERO-PRO and RoboCasa365. (c) It reaches a 58.9\% two-domain average, surpassing all consolidated baselines and both specialists.}
  \label{fig:teaser}
  \vspace{-3mm}
\end{figure}

Building general-purpose robots has long been a central goal, and agentic robotics has recently emerged as a promising path toward it~\citep{codeaspolicies,capx,harnessvla}. In this paradigm, a multimodal \emph{planner} interprets the instruction, decomposes the task, and acts by calling \emph{tools}: callable functions that either perceive the scene (e.g., segmenting an object or locating it in 3D) or invoke a frozen low-level \emph{executor}, such as a motion primitive or a vision-language-action (VLA) policy. After each call, the tool returns feedback from the environment, such as the execution status and updated camera views. This loop closely resembles how recent coding agents work. Because low-level executors handle embodiment-specific control, the planner only reasons and calls tools rather than directly outputting motor commands. As with coding agents, achieving general capability then comes down to obtaining a capable planner.

However, most existing planners are proprietary models prompted through APIs~\citep{capx,openrua,harnessvla}, whose per-call cost and latency accumulate over long closed-loop episodes and rule out on-device deployment. To remove this dependence, recent post-training methods~\citep{agenticroborl,guava,captain} train open-weight models with supervised fine-tuning (SFT) followed by multi-turn reinforcement learning (RL). Each resulting planner is trained for a single \emph{domain}, i.e., one robot environment together with its executors and tools. Because tools differ across domains in names, arguments, and feedback formats, every domain requires its own planner, and serving $N$ domains means deploying $N$ planners (Figure~\ref{fig:teaser}a). This raises a natural question: \textbf{\emph{Can we train a single open-weight planner to operate across multiple domains, as a step toward general-purpose robots?}}

A natural approach is to \emph{consolidate} specialist planners across domains into a single \emph{generalist planner}. Large language model (LLM) post-training offers four common consolidation methods: weight merging~\citep{modelsoups,taskarithmetic,mopd}, off-policy specialist SFT~\citep{mopd,openmopd,kdrl}, multi-domain online RL~\citep{nemotroncascade2,mopd,openmopd}, and multi-teacher on-policy distillation (MOPD)~\citep{mimov2flash,nemotroncascade2,mopd,openmopd}. Agentic robotics, however, differs fundamentally from the language and software domains where these methods were developed, so conclusions drawn there do not transfer directly. First, domains differ in robot embodiment, camera views, and low-level tools. Second, the planner acts over dozens of turns from multi-view camera images, repeatedly recovering when stochastic physical execution slips or fails. Third, a physical task often fails because a single step selects the wrong tool or a slightly inaccurate spatial parameter, yet the episode outcome reveals only success or failure, not where the error occurred. Under these conditions, it remains unclear how these consolidation methods should be adapted to agentic robotics and which of them actually work.

To answer this question, we compare the four methods on the same Qwen3.5-9B backbone~\citep{qwen35}, with the three training-based methods starting from a shared cross-domain SFT initialization under a matched consolidation rollout budget. To include MOPD in this comparison, we adapt it to agentic robotics as \textbf{Robo-MOPD} (Figure~\ref{fig:overview}). In Robo-MOPD, the generalist student runs closed-loop rollouts in simulation, and the matching frozen specialist scores each token of the student's reasoning and tool call via token-level reverse KL, correcting errors such as a wrong tool or an inaccurate coordinate at the step where they occur. This same scoring scheme works across all robots because, even when embodiments and tools differ, every turn shares the same format: camera images and tool feedback go in, and reasoning plus a tool call come out. Across all training-based methods, we drop prior-turn reasoning from the context~\citep{stateopd}, subsample a random subset of turns per episode~\citep{agenticroborl}, and balance the loss across episode lengths~\citep{minillm,mopd,kdrl} and domains~\citep{openmopd} to keep training efficient and balanced over long, multi-view episodes (Section~\ref{sec:method}). All low-level executors~\citep{pi05,rldx} remain frozen throughout. At test time, a single generalist planner operates robots across all domains without any specialist.

We evaluate on tabletop LIBERO-PRO~\citep{libero,liberopro} and mobile-kitchen RoboCasa365~\citep{robocasa,robocasa365} and obtain three findings. \uline{First, a single generalist planner can outperform the specialists.} Robo-MOPD reaches 72.9\% on LIBERO-PRO and 44.8\% on RoboCasa365, the only method to surpass both 9B specialists, achieving a 58.9\% average while all baseline methods fall below the specialists' 56.9\% (Figure~\ref{fig:teaser}b,c and Table~\ref{tab:main-consolidation}). \uline{Second, the gap concentrates on long-horizon tasks.} On the long-horizon LIBERO-10 suite, neither off-policy SFT nor online RL improves over the shared SFT initialization, while Robo-MOPD leads online RL by 13.0 percentage points (Figure~\ref{fig:liberopro-suites}). \uline{Third, Robo-MOPD changes how the planner fails.} On LIBERO-PRO, budget-exhaustion failures drop from 31.6\% under online RL to 20.3\%, while voluntary \texttt{finish()} calls on unsolved episodes rise from 2.6\% to 6.8\%: on hard tasks, Robo-MOPD stops voluntarily after failed attempts rather than repeating actions until the 100-turn limit like the baselines (Section~\ref{sec:exp-analysis} and Figure~\ref{fig:suppl-turn-timeout}).

Our contributions are as follows:
\begin{itemize}
    \item \textbf{Systematic comparison.} We present the first systematic comparison of four consolidation methods for agentic robot planners: weight merging, off-policy specialist SFT, multi-domain online RL, and MOPD.
    \item \textbf{Robo-MOPD.} To our knowledge, we are the first to bring MOPD to agentic robotics, adapting it to closed-loop multimodal planning so that a single generalist planner spans incompatible robot domains toward general-purpose robots.
    \item \textbf{Findings and performance.} Consolidation works best when it combines on-policy rollouts with token-level supervision, and its advantage is largest on long-horizon tasks. A single 9B generalist planner surpasses both 9B domain specialists.
\end{itemize}

\section{Related Work}
\label{sec:related}

\paragraph{Agentic Robot Harnesses.}
In agentic robotics, a domain's tool interface and closed-loop execution wrapper is commonly referred to as a \emph{robot harness}. Language and vision-language models guide robots by sequencing learned skills~\citep{saycan}, generating control programs~\citep{codeaspolicies}, specifying spatial or keypoint constraints~\citep{voxposer,rekep}, or steering motor policies in hierarchies~\citep{hirobot}. Recent agentic systems invoke VLA skills or analytic primitives in execution loops~\citep{agenticrobot,harnessvla,physicalagency,vlasastools,pyrua}, synthesize control code~\citep{capx}, expose modular perception and workspace interfaces~\citep{spark,openrua,unirobotagent}, or couple code skills with video and action generation~\citep{aw0}. Other harnesses add spatial measurement~\citep{roboharnessk1}, action rehearsal~\citep{worldactionagent}, safe motion routing~\citep{safeharness}, task memories~\citep{baton,roboharness}, video keyframes~\citep{rvicl}, or distilled playbooks~\citep{rhd}. Related systems build reusable skill libraries from exploration~\citep{aspire,rats,racap}, evolve harnesses online~\citep{dynaharness,roborsi}, or optimize primitives before deployment~\citep{rho}. Rather than expanding external tools, memories, or skill libraries around a fixed model, we keep each harness and its executors fixed and consolidate planner weights across domains.

\paragraph{Post-Training Robot Planners.}
RL improves embodied planners for spatial reasoning, visual state modeling, and mobile manipulation~\citep{embodiedr,vagen,real}. Agent RL systems decouple rollout execution or asynchronous optimization from policy updates~\citep{agentlightning,sao}, though joint RL across coding harnesses adapts to interface conventions rather than transferring skills~\citep{multiharnessrl}. Within robot harnesses, recent methods fine-tune compact students on teacher or recovery trajectories~\citep{guava,roboharnessk1,worldactionagent}, internalize validated interaction logs into an embodied planner~\citep{sensenovaroborsi}, or apply group relative policy optimization~\citep{deepseekmath} over code and primitive interfaces~\citep{capx,pyrua,guava}. CaPtain~\citep{captain} trains a 4B planner to write control code over perception and motion primitives via SFT and streaming multi-turn RL~\citep{sao} with a physical critic over privileged simulator state. AgenticRoboRL~\citep{agenticroborl} pairs a serving proxy with harness adapters to run demonstration SFT without external memories and multi-turn outcome RL across fixed VLA and code harnesses. These methods train one specialist per harness or domain, whereas Robo-MOPD consolidates frozen specialists across harnesses, embodiments, and task suites into one generalist planner.

\paragraph{On-Policy Distillation and Specialist Consolidation.}
Building on interactive imitation learning~\citep{dagger}, on-policy distillation trains a student on its own rollouts under dense token divergence supervision~\citep{minillm,gkd,thinkingmachinesopd,rethinkingopd}. Recent studies gate teacher guidance on failed rollouts where outcome RL~\citep{dapo} lacks positive signal~\citep{kdrl,rstg}, fuse bounded distillation advantages with RL~\citep{safopd}, or supervise multi-turn text agents under one teacher~\citep{turnopd,memopd,stateopd}. Across multiple domains, parameter merging combines weights or task vectors~\citep{modelsoups,taskarithmetic}, whereas MOPD routes student rollouts to matching specialists for token supervision in language domains~\citep{mimov2flash,mopd,openmopd,nemotroncascade2}. In robotics, scaling end-to-end VLA policies~\citep{openvla,pi0,pi05,rlinfvla} across robots requires large cross-embodiment datasets~\citep{openvla,pi0,xvla}, and fine-tuned policies can forget prior skills on new tasks or embodiments~\citep{xvla,vlaopd}. Recent work adapts or merges VLAs via task-routed masked adapters~\citep{mergevla}, embodiment-specific soft prompts~\citep{xvla}, dual-teacher distillation~\citep{dualvla}, offline specialist or residual RL rollouts~\citep{rldg,pld}, or single-teacher on-policy distillation~\citep{vlaopd,wamopd}. Without test-time routing, direct VLA weight merging collapses~\citep{mergevla}, and concurrent multi-teacher flow distillation~\citep{flowvlamopd} across distinct continuous action spaces still yields consolidated policies that trail individual specialists. Whereas language MOPD shares one text interface and VLA consolidation updates low-level action generators, Robo-MOPD keeps perception tools, motion primitives, and VLA controllers~\citep{pi05,rldx} frozen and consolidates high-level planners across distinct harnesses.

\section{Method}
\label{sec:method}

\begin{figure*}[t]
  \centering
  \includegraphics[width=\linewidth]{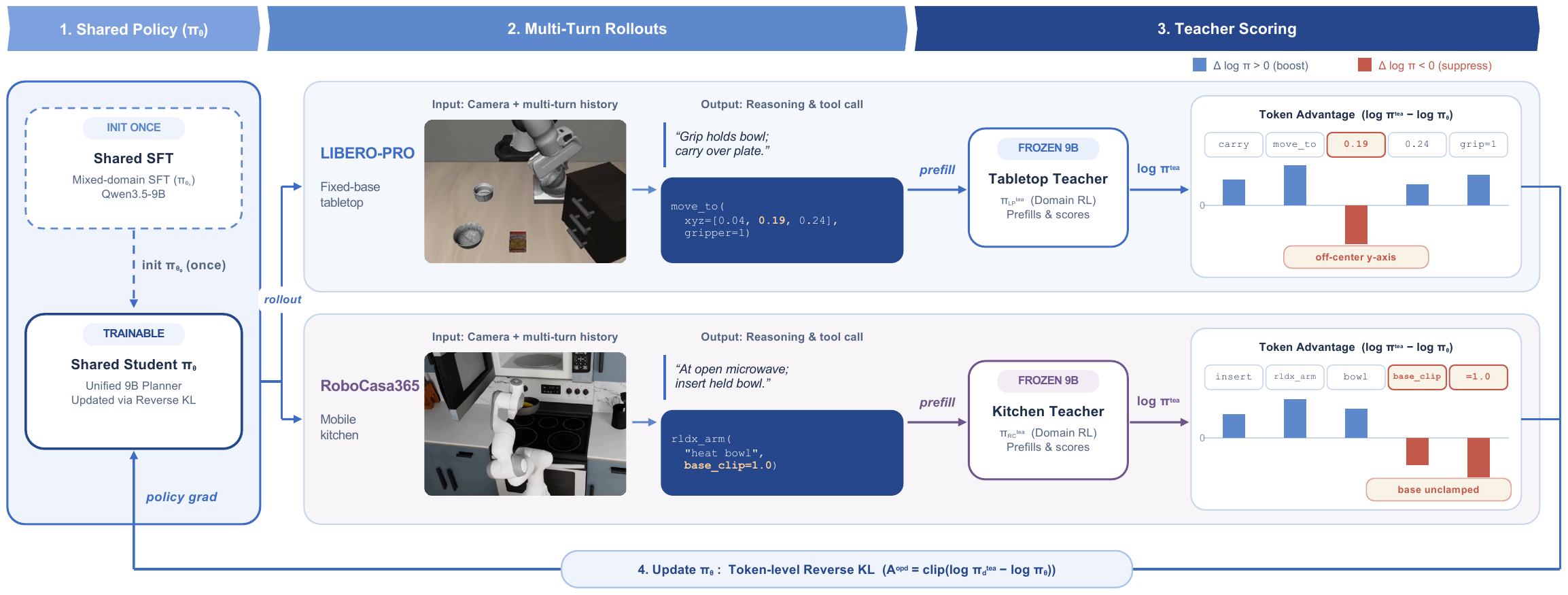}
  \vspace{-6mm}
  \caption{\textbf{Overview of Robo-MOPD.} (1) Shared SFT initializes the unified 9B student planner $\pi_\theta$ once ($\pi_{\theta_0}$). (2) In each round, $\pi_\theta$ runs closed-loop rollouts across LIBERO-PRO and RoboCasa365, mapping camera views and history to reasoning and a tool call. (3) The matching frozen specialist ($\pi_{\mathrm{LP}}^{\mathrm{tea}}$ or $\pi_{\mathrm{RC}}^{\mathrm{tea}}$) scores the student's tokens, boosting valid decisions and suppressing flawed arguments such as an off-center $y$-coordinate (\texttt{0.19}) or unrestricted mobile base motion (\texttt{base\_clip=1.0}). (4) Clipped token advantages $A^{\mathrm{opd}}$ update $\pi_\theta$ via token-level reverse KL.}
  \label{fig:overview}
  \vspace{-4mm}
\end{figure*}

\subsection{Multi-Turn Robot Harness Setup}
\label{sec:problem-setup}

We consider a set of robot manipulation domains $\mathcal{D}$, where each domain $d \in \mathcal{D}$ pairs a distinct robot embodiment, workspace, and camera layout with its own software harness~\citep{harnessvla}. Each task instance in domain $d$ specifies a language instruction and an initial simulator state. Rather than predicting continuous motor commands, a vision-language planner $\pi_\theta$ interacts with domain $d$ over closed-loop turns by calling harness tools, including spatial and camera queries, scripted motions, and frozen vision-language-action (VLA) policies~\citep{codeaspolicies,harnessvla}.

Following recent agent post-training frameworks~\citep{agentlightning,legorl}, one decision turn corresponds to one planner call. At turn $m \in \{1, \dots, M_e\}$ of episode $e$, the harness constructs a multimodal prompt context $x_{e,m}$ from the task instruction, prior tool calls and execution feedback, and multi-view camera images while omitting reasoning text from earlier turns. Conditioned on $x_{e,m}$, the planner generates a token sequence $y_{e,m} = (y_{e,m,1}, \dots, y_{e,m,n_{e,m}})$ containing reasoning text and a tool call according to $\pi_\theta(y_{e,m} \mid x_{e,m}) = \allowbreak \prod_{t=1}^{n_{e,m}} \pi_\theta(y_{e,m,t} \mid x_{e,m}, y_{e,m,<t})$. Executing the requested tool in simulation returns updated feedback and camera views to form $x_{e,m+1}$, and this cycle repeats until the planner calls the finish tool or hits a turn cap $M_{\max}$. This loop yields a trajectory $\tau_e = ((x_{e,1}, y_{e,1}), \allowbreak \dots, \allowbreak (x_{e,M_e}, y_{e,M_e}))$ of variable turn length $M_e$, where neither turn index $m$ nor token index $t$ corresponds to a motor control step.

Given frozen specialist planners $\{\pi_d^{\mathrm{tea}}\}_{d \in \mathcal{D}}$ post-trained independently per domain~\citep{agenticroborl}, we introduce \textbf{Robo-MOPD} to consolidate them into one generalist planner $\pi_\theta$ across $\mathcal{D}$. At test time, $\pi_\theta$ operates without external memory (e.g., prior task logs), specialist routing, or separate checkpoints while keeping all robot tools and VLA policies fixed. As shown in Figure~\ref{fig:overview}, training combines Stage~1 cross-domain supervised initialization with Stage~2 iterative on-policy distillation~\citep{mopd,openmopd} on the student's own simulator rollouts.

\subsection{Cross-Domain Robot Planner Initialization}
\label{sec:pipeline}

Post-training consolidation requires a shared initial planner $\pi_{\theta_0}$ that aligns with the specialists~\citep{rethinkingopd,mopd,openmopd} and emits valid tool calls across robot harnesses so simulator rollouts reach informative task states. In practice, source demonstrations are collected using a planner with external memory via file access to prior task logs on the same task~\citep{harnessvla}, though such memory is unavailable at evaluation time. Following prior supervision adaptation~\citep{agenticroborl}, we strip task log reads and outputs and rewrite reasoning text so decisions rely only on camera views and tool feedback. We then replay each adapted trajectory through its native student harness to reconstruct exact prompt contexts, keeping any turn rejected by an automated language model judge in the interaction history for state continuity while masking its tokens from the loss. Pooling verified turns $(x_i, y_i)$ across all domains $d \in \mathcal{D}$, we train $\pi_{\theta_0}$ by minimizing token-level cross-entropy
\begin{equation}
  \mathcal{L}_{\mathrm{SFT}}(\theta)=
-\frac{
\displaystyle
\sum_{(x_i, y_i) \in \mathcal{B}_{\mathrm{SFT}}}
\sum_{t=1}^{n_i}
\log \pi_\theta(y_{i,t} \mid x_i, y_{i,<t})
}{
\displaystyle
\sum_{(x_i, y_i) \in \mathcal{B}_{\mathrm{SFT}}} n_i
}
\label{eq:sft-loss}
\end{equation}
over turn minibatches $\mathcal{B}_{\mathrm{SFT}}$, where turn $i$ has $n_i$ response tokens and masked prompt context $x_i$.

\subsection{Closed-Loop Robot Rollouts and Specialist Routing}
\label{sec:rollout-routing}

Starting from $\pi_{\theta_0}$, Stage~2 runs over iterative online rounds. At the start of each round, we freeze the current student robot planner as the rollout policy $\pi_{\mathrm{old}}$ and sample manipulation tasks in every domain $d \in \mathcal{D}$. For each sampled task and initial scene state, $\pi_{\mathrm{old}}$ executes $K$ independent multi-turn episodes in the simulator to form a rollout group $\mathcal{G}$, where each episode $e \in \mathcal{G}$ receives a binary task completion indicator $R_e \in \{0, 1\}$ from the simulator goal check. This indicator is used only for group filtering and does not enter the distillation loss. As in dynamic group filtering for policy optimization~\citep{dapo}, we discard groups in which all $K$ attempts already succeed ($\sum_{e \in \mathcal{G}} R_e = K$) so updates concentrate on unmastered scene configurations. Whereas group filtering from binary rewards alone also drops groups with zero successes~\citep{dapo}, we retain all episodes from both mixed groups ($0 < \sum_{e \in \mathcal{G}} R_e < K$) and unsolved groups ($\sum_{e \in \mathcal{G}} R_e = 0$) in the active episode set $\mathcal{E}_d$ of robot domain $d$, since the specialist still provides token-level supervision when every attempt in a group fails~\citep{kdrl,rstg}.

To bound compute and teacher scoring cost over long manipulation horizons, we form a chronological turn index set $\mathcal{M}_e \subseteq \{1, \dots, M_e\}$ for each active episode $e \in \mathcal{E}_d$, uniformly subsampling $L$ indices without replacement when $M_e > L$ and retaining all $M_e$ indices otherwise~\citep{agenticroborl}. Following domain routing in multi-teacher distillation~\citep{mopd,openmopd}, each selected turn $(x_{e,m}, y_{e,m})$ with $m \in \mathcal{M}_e$ is routed exclusively to the matching specialist robot planner $\pi_d^{\mathrm{tea}}$ rather than averaging predictions across specialists trained on incompatible workspaces, camera setups, and harnesses. For each routed turn, one forward pass of $\pi_d^{\mathrm{tea}}$ evaluates $\log \pi_d^{\mathrm{tea}}(y_{e,m,t} \mid s_{e,m,t})$ across all student response tokens $t \in \{1, \dots, n_{e,m}\}$ conditioned on the multimodal prefix $s_{e,m,t} = (x_{e,m}, y_{e,m,<t})$ observed in simulation, including camera observations after imperfect robot actions.

\subsection{On-Policy Robot Planner Distillation}
\label{sec:mopd-objective}

Following the policy gradient formulation of reverse Kullback--Leibler distillation~\citep{minillm,thinkingmachinesopd,mopd}, we supervise each student response token $y_{e,m,t}$ across reasoning text and robot tool calls using the immediate log probability gap between the routed specialist $\pi_d^{\mathrm{tea}}$ and the current student $\pi_\theta$. As illustrated in Figure~\ref{fig:overview}, scoring individual tokens rewards valid state reasoning and tool selection while suppressing an off-center coordinate or unrestricted mobile base motion during arm insertion within the same turn. Because token log probability gaps have heavy tails, we clip this token advantage to $[C_{\min}, C_{\max}]$ with $C_{\min} < 0 < C_{\max}$ to keep extreme tokens from dominating updates~\citep{mopd,safopd},
\begin{equation}
\begin{aligned}
A_{e,m,t}^{\mathrm{opd}}
&= \operatorname{sg}\!\Bigl[\mathrm{clip}\!\bigl(
\log \pi_d^{\mathrm{tea}}(y_{e,m,t} \mid s_{e,m,t}) \\
&\quad - \log \pi_\theta(y_{e,m,t} \mid s_{e,m,t}),\,
C_{\min},\,
C_{\max}
\bigr)\Bigr],
\end{aligned}
\label{eq:opd-advantage}
\end{equation}
where $\operatorname{sg}[\cdot]$ denotes the stop-gradient operator so that $A_{e,m,t}^{\mathrm{opd}}$ is evaluated as a detached weight under the current planner parameters $\theta$ at each update step~\citep{openmopd}, while $\pi_{\mathrm{old}}$ remains fixed for the round. Within each round, selected planning turns from all robot domains are shuffled into minibatches for one pass of gradient updates. With the token importance ratio
\begin{equation}
\rho_{e,m,t}(\theta)
=
\frac{\pi_\theta(y_{e,m,t} \mid s_{e,m,t})}{\pi_{\mathrm{old}}(y_{e,m,t} \mid s_{e,m,t})},
\label{eq:importance-ratio}
\end{equation}
substituting $A_{e,m,t}^{\mathrm{opd}}$ into the clipped policy gradient surrogate~\citep{ppo,dapo} yields the token distillation loss
\begin{equation}
\begin{aligned}
\ell_{e,m,t}^{\mathrm{opd}}(\theta)
&= \max\!\Bigl(
-A_{e,m,t}^{\mathrm{opd}}\,\rho_{e,m,t}(\theta), \\
&\quad -A_{e,m,t}^{\mathrm{opd}}\,
\mathrm{clip}\!\bigl(\rho_{e,m,t}(\theta),\, 1 - \epsilon_-,\, 1 + \epsilon_+\bigr)
\Bigr),
\end{aligned}
\label{eq:token-ppo-loss}
\end{equation}
where $\epsilon_-$ and $\epsilon_+$ denote the lower and upper clipping bounds~\citep{dapo}, and whenever $A_{e,m,t}^{\mathrm{opd}} < 0$ we cap $\ell_{e,m,t}^{\mathrm{opd}}(\theta)$ at $-c\,A_{e,m,t}^{\mathrm{opd}}$ for $c > 1$, removing gradients on tokens with negative advantage once $\rho_{e,m,t}(\theta) > c$~\citep{dualclipppo}.

In closed-loop robot manipulation, failed episodes that repeatedly query cameras and retry tools until the turn limit produce far more tokens than concise successful episodes~\citep{turnopd}, while domains with higher solve rates discard more solved groups and leave unequal active episode counts. Using sequence length normalization~\citep{minillm,mopd,kdrl} with equal weight across robot domains~\citep{openmopd}, we define the round objective by averaging token losses within each active episode and across domains,
\begin{equation}
\begin{aligned}
\mathcal{L}_{\mathrm{MOPD}}(\theta)
&= \frac{1}{|\mathcal{D}|}
\sum_{d \in \mathcal{D}}
\frac{1}{|\mathcal{E}_d|}
\sum_{e \in \mathcal{E}_d} \\
&\quad
\frac{1}{T_e}
\sum_{m \in \mathcal{M}_e}
\sum_{t=1}^{n_{e,m}}
\ell_{e,m,t}^{\mathrm{opd}}(\theta),
\end{aligned}
\label{eq:mopd-total-loss}
\end{equation}
where $T_e = \sum_{m \in \mathcal{M}_e} n_{e,m}$ is the total number of selected response tokens in episode $e$. Weighting each selected token by $1 / (|\mathcal{D}|\,|\mathcal{E}_d|\,T_e)$ makes each minibatch update an unbiased estimate of Eq.~\eqref{eq:mopd-total-loss} balanced across episodes and domains. Under this shared $\pi_{\theta_0}$, turn subsampling $\mathcal{M}_e$, and normalization, multi-domain online RL replaces $A_{e,m,t}^{\mathrm{opd}}$ with group outcome advantages, while off-policy specialist SFT minimizes cross-entropy on specialist rollouts, and weight merging combines specialist parameters directly (Section~\ref{sec:exp-setup} and Appendix~\ref{sec:suppl-objectives}).

\section{Experiments}
\label{sec:experiments}

\subsection{Experimental Setup}
\label{sec:exp-setup}

\paragraph{Benchmarks and evaluation protocol.}
We evaluate consolidation across two manipulation benchmarks that differ in robot embodiment, camera layout, and callable tool interfaces~\citep{harnessvla} (Table~\ref{tab:suppl-harness-apis}). For tabletop manipulation, LIBERO-PRO~\citep{libero,liberopro} controls a Franka Panda arm through spatial and camera queries, analytic arm and gripper motions, and a frozen $\pi_{0.5}$ VLA policy~\citep{pi05,rlinf}. Following the evaluation protocol of prior harness planners~\citep{harnessvla}, we evaluate 80 tasks across eight suites spanning Spatial, Object, Goal, and LIBERO-10 under instruction redirection and position swap perturbations on 10 held-out scene states per task (800 evaluation episodes). For mobile kitchen manipulation, RoboCasa365~\citep{robocasa,robocasa365} controls a mobile manipulator across counters, cabinets, doors, and appliances through navigation and manipulation tools paired with a frozen RLDX-1 VLA controller~\citep{rldx}. Using the mobile kitchen harness and 50-task target split from prior work~\citep{robocasa365,harnessvla}, we evaluate held-out seeds~1--5 per task (250 evaluation episodes) across 18 atomic tasks, 16 composite tasks seen during RLDX-1 controller pretraining, and 16 composite tasks unseen during RLDX-1 pretraining. Both Stage~1 adaptation and Stage~2 rollouts use training scene states ($\ge 11$) disjoint from all evaluation scene states. At evaluation time, each consolidated planner operates both native tool harnesses from one 9B parameter set without prior task logs, environment resets, or external specialist routing, decoding with temperature $1.0$, $\text{top-}p = 0.95$, $\text{top-}k = 20$, presence penalty $1.5$, and turn limit $M_{\max} = 100$. We report single-run simulator completion rates and the mean (Avg.\@) of LIBERO-PRO Overall and RoboCasa365 Overall.

\paragraph{Planners and shared initialization.}
Both specialists and the generalist student use Qwen3.5-9B~\citep{qwen35}, freezing the vision encoder while updating the language backbone and visual merger with all robot tools and VLA controllers fixed. Each frozen specialist $\pi_d^{\mathrm{tea}}$ is trained independently from Base Qwen3.5-9B via single-domain SFT and six rounds of online RL~\citep{agenticroborl}. Adapting Qwen3.8-27B~\citep{qwen38} demonstrations on training states as in Section~\ref{sec:pipeline} yields 55{,}781 verified turns across both domains, and minimizing Eq.~\eqref{eq:sft-loss} for one epoch produces Shared SFT ($\pi_{\theta_0}$), the initial checkpoint for all training-based methods.

\paragraph{Shared online training protocol.}
All online consolidation methods start from $\pi_{\theta_0}$ and run for eight rounds under matched Stage~2 consolidation rollout budgets, identical sampled scenes, and shared optimizer settings, using Round~8 for all endpoint comparisons. In every round, the rollout policy $\pi_{\mathrm{old}}$ executes 32 groups of $K = 6$ episodes per domain at temperature $1.0$ and $M_{\max} = 100$ turns on training states 11--44 without replacement across rounds. This schedule yields 192 episodes per domain and 384 per round, totaling 3{,}072 episodes across Rounds~1--8. Over each method's retained episode set $\mathcal{E}_d$, we uniformly subsample at most $L = 16$ planning turns per episode in chronological order and apply $1 / (|\mathcal{D}|\,|\mathcal{E}_d|\,T_e)$ in Eq.~\eqref{eq:mopd-total-loss} for $1 / T_e$ token normalization~\citep{minillm,mopd,kdrl} and $1 / |\mathcal{D}| = 0.5$ domain weighting~\citep{openmopd}. All online methods share PPO ratio clipping $(\epsilon_-, \epsilon_+) = (0.20, 0.28)$~\citep{dapo}, cap $c = 3.0$ on negative advantages~\citep{dualclipppo}, one pass of updates at global batch size 128 turns, constant learning rate $10^{-6}$, and $\ell_2$ gradient clipping at $1.0$.

\paragraph{Baselines.}
We compare \textbf{Robo-MOPD} against four consolidation baselines on Qwen3.5-9B, with full formulations in Appendix~\ref{sec:suppl-objectives}. (1) \textbf{Weight Averaging}~\citep{modelsoups} ($\alpha = 0.5$) uniformly averages the weights of both 9B specialists. (2) \textbf{Task Arithmetic}~\citep{taskarithmetic} ($\alpha = 1.0$) adds both specialist task vectors relative to Base Qwen3.5-9B, whereas setting $\alpha = 0.5$ reduces to Weight Averaging for $|\mathcal{D}| = 2$. (3) \textbf{Off-Policy Specialist SFT}~\citep{mopd,openmopd,rldg} rolls out each frozen 9B specialist across the identical 3{,}072-episode Stage~2 scene schedule without task logs and discards failed episodes~\citep{kdrl,openmopd,rldg} so cross-entropy does not imitate unsuccessful actions. Starting from $\pi_{\theta_0}$, it runs one pass of forward token cross-entropy under the same $L = 16$ turn subsampling and $1 / (|\mathcal{D}|\,|\mathcal{E}_d|\,T_e)$ normalization, and we evaluate the final checkpoint to match the Round~8 endpoint protocol without intermediate selection~\citep{flowvlamopd}. (4) \textbf{Multi-Domain Online RL}~\citep{mopd,openmopd} follows the same eight-round protocol from $\pi_{\theta_0}$, replacing specialist reverse KL with group leave-one-out advantages~\citep{rloo,agenticroborl} from binary completion rewards~\citep{dapo}.

\providecommand{\robocmark}{\textcolor[rgb]{0.11,0.47,0.22}{\checkmark}}
\providecommand{\roboxmark}{\textcolor[rgb]{0.70,0.09,0.17}{$\times$}}
\providecommand{\robonamark}{\textcolor[rgb]{0.40,0.40,0.40}{--}}
\begin{table*}[t]
\centering
\small
\setlength{\tabcolsep}{2.9pt}
\begin{tabular}{@{}l ccc ccccc cccc c@{}}
\toprule
& \multicolumn{3}{c}{Consolidation Setup} & \multicolumn{5}{c}{LIBERO-PRO (800 eps)} & \multicolumn{4}{c}{RoboCasa365 (250 eps)} & \\
\cmidrule(lr){2-4} \cmidrule(lr){5-9} \cmidrule(lr){10-13}
Method &
\begin{tabular}[b]{@{}c@{}}Single\\[-1.5pt]$1{\times}\text{9B}$\end{tabular} &
\begin{tabular}[b]{@{}c@{}}On-\\[-1.5pt]Policy\end{tabular} &
\begin{tabular}[b]{@{}c@{}}Token\\[-1.5pt]Sup.\end{tabular} &
Spatial & Object & Goal & L10 & Overall & Atomic & Seen & Unseen & Overall & Avg. \\
\midrule
\multicolumn{14}{@{}l}{\textit{Reference policies}} \\
Base Qwen3.5-9B & \robocmark & \robonamark & \robonamark & 19.5 & 17.0 & 29.5 & 6.5 & 18.1 & 46.7 & 15.0 & 1.2 & 22.0 & 20.1 \\
Shared SFT ($\pi_{\theta_0}$) & \robocmark & \roboxmark & \robocmark & 79.5 & 88.0 & 51.0 & 31.0 & 62.4 & 71.1 & 23.8 & 10.0 & 36.4 & 49.4 \\
Domain Specialists ($2{\times}\text{9B}$) & \roboxmark & \robonamark & \robonamark & 91.0 & 95.5 & 57.5 & 40.0 & 71.0 & 77.8 & 30.0 & 16.2 & 42.8 & 56.9 \\
\midrule
\multicolumn{14}{@{}l}{\textit{Consolidated generalist 9B planners}} \\
Weight Averaging ($\alpha = 0.5$) & \robocmark & \robonamark & \roboxmark & 80.0 & 81.0 & 52.0 & 21.0 & 58.5 & 74.4 & 20.0 & 8.8 & 36.0 & 47.3 \\
Task Arithmetic ($\alpha = 1.0$) & \robocmark & \robonamark & \roboxmark & 25.0 & 48.0 & 23.5 & 4.0 & 25.1 & 50.0 & 11.2 & 5.0 & 23.2 & 24.2 \\
Off-Policy Specialist SFT & \robocmark & \roboxmark & \robocmark & 80.0 & 92.5 & 56.0 & 29.0 & 64.4 & 73.3 & 28.8 & 13.8 & 40.0 & 52.2 \\
Multi-Domain Online RL & \robocmark & \robocmark & \roboxmark & 81.5 & 95.5 & 57.5 & 28.5 & 65.8 & \textbf{77.8} & \textbf{37.5} & \textbf{20.0} & \textbf{46.4} & 56.1 \\
\textbf{Robo-MOPD (Ours)} & \robocmark & \robocmark & \robocmark & \textbf{88.0} & \textbf{96.0} & \textbf{66.0} & \textbf{41.5} & \textbf{72.9} & 76.7 & \textbf{37.5} & 16.2 & 44.8 & \textbf{58.9} \\
\midrule
$\Delta$ \textit{(Ours $-$ Domain Specialists)} & & & & \textit{$-$3.0} & \textit{$+$0.5} & \textit{$+$8.5} & \textit{$+$1.5} & \textit{$+$1.9} & \textit{$-$1.1} & \textit{$+$7.5} & \textit{$+$0.0} & \textit{$+$2.0} & \textit{$+$2.0} \\
\bottomrule
\end{tabular}
\vspace{-2mm}
\caption{Reference policies and Round~8 consolidation success rates (\%) on LIBERO-PRO (200 episodes per task family across Spatial, Object, Goal, and LIBERO-10, totaling 800 episodes) and RoboCasa365 (90 Atomic, 80 Seen composite, and 80 Unseen composite episodes defined by RLDX-1 controller pretraining, totaling 250 episodes), along with each method's properties across single 9B deployment ($1{\times}\text{9B}$), closed-loop student rollouts (On-Policy), and per-token supervision (Token Sup.\@). Consolidated planners run from one 9B checkpoint without prior task logs or specialist routing. Domain Specialists evaluates each separate 9B specialist on its native benchmark, and Avg.\ is the mean of the reported LIBERO-PRO Overall and RoboCasa365 Overall columns. Bold marks the best consolidated 9B planner in each column.}
\label{tab:main-consolidation}
\vspace{-4mm}
\end{table*}

\begin{figure*}[t]
\centering
\includegraphics[width=0.88\textwidth]{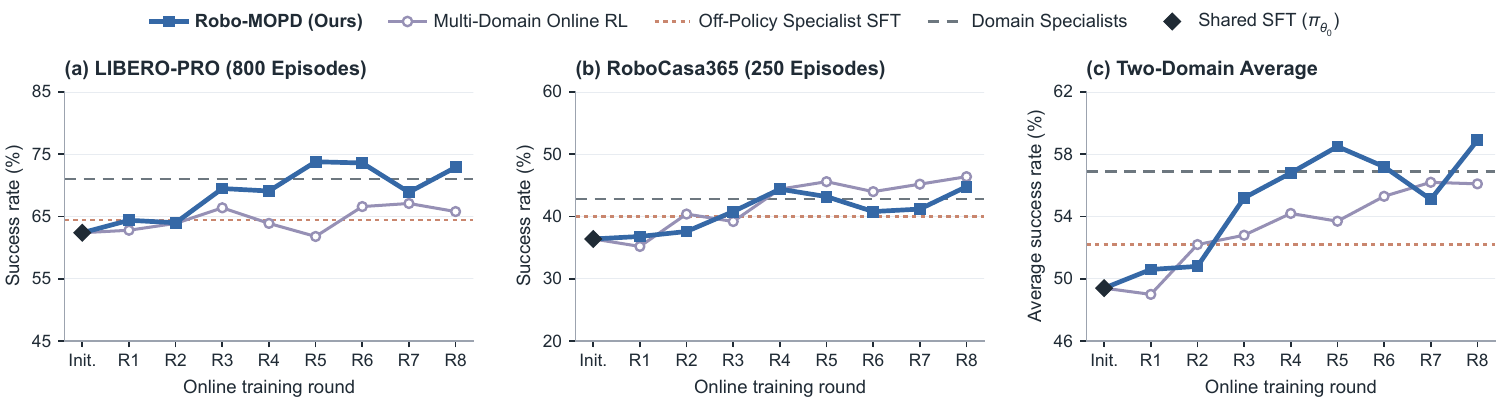}
\vspace{-1mm}
\caption{Evaluation success rates across Rounds~1--8 (R1--R8) on (a) LIBERO-PRO (800 episodes), (b) RoboCasa365 (250 episodes), and (c) their two-domain average. Diamonds at Init.\ mark Shared SFT ($\pi_{\theta_0}$), and horizontal dashed lines mark Domain Specialists ($\pi_d^{\mathrm{tea}}$) and Off-Policy Specialist SFT. Robo-MOPD improves both domains over $\pi_{\theta_0}$ and surpasses the pair of separate 9B specialists with a single 9B planner.}
\label{fig:main-curves}
\vspace{-4mm}
\end{figure*}

\subsection{Main Consolidation Results}
\label{sec:exp-main}

Table~\ref{tab:main-consolidation} reports Round~8 consolidation performance across LIBERO-PRO and RoboCasa365, and Figure~\ref{fig:main-curves} plots evaluation trajectories across Rounds 1--8, with per-round scores and per-suite splits in Tables~\ref{tab:suppl-all-rounds}--\ref{tab:liberopro-suites}.

\paragraph{Offline initialization, parameter merging, and off-policy specialist SFT.}
Stage~1 initialization ($\pi_{\theta_0}$) raises Base Qwen3.5-9B from 18.1\% to 62.4\% on LIBERO-PRO and from 22.0\% to 36.4\% on RoboCasa365, yet remains 8.6 and 6.4 points behind the separate 9B specialists (71.0\% and 42.8\%, 49.4\% vs.\ 56.9\% Avg.), reflecting the gap between offline adapted traces and closed-loop execution. Merging the two specialists in parameter space degrades multi-turn control. Weight Averaging ($\alpha = 0.5$) drops to 58.5\% on LIBERO-PRO and 36.0\% on RoboCasa365 (47.3\% Avg.), falling below $\pi_{\theta_0}$ and losing 19.0 points to the tabletop specialist on LIBERO-10 (21.0\% vs.\ 40.0\%). Because each specialist underwent full single-domain supervised and reinforcement post-training from Base Qwen3.5-9B, adding unscaled task vectors in Task Arithmetic ($\alpha = 1.0$) compounds both parameter shifts and drops performance to 25.1\% and 23.2\% (24.2\% Avg.), while tuning $\alpha$ to $0.5$ coincides with Weight Averaging (47.3\% Avg.). Fine-tuning $\pi_{\theta_0}$ on successful specialist rollouts in Off-Policy Specialist SFT improves both benchmarks to 64.4\% on LIBERO-PRO and 40.0\% on RoboCasa365 (52.2\% Avg., +2.8 points over $\pi_{\theta_0}$), yet still falls 4.7 points short of the separate specialists and trails both online methods.

\paragraph{Comparison with multi-domain online RL and effect of specialist teacher strength.}
Comparing Figure~\ref{fig:main-curves}a and Figure~\ref{fig:main-curves}b shows how online distillation and outcome reinforcement learning depend on the capability of each domain specialist. On RoboCasa365 (Figure~\ref{fig:main-curves}b), Multi-Domain Online RL reaches 46.4\% at Round~8, running slightly above Robo-MOPD at 44.8\%. Because Robo-MOPD learns strictly from routed token probabilities without terminal task rewards, its kitchen gains are constrained by the relatively modest 42.8\% RoboCasa365 specialist, whereas optimizing simulator completion rewards directly allows Multi-Domain Online RL to explore beyond that weaker teacher. Even with this modest teacher, Robo-MOPD still improves over $\pi_{\theta_0}$ by +8.4 points and edges past the 42.8\% kitchen specialist by +2.0 points.

\begin{figure*}[t]
\centering
\includegraphics[width=0.88\textwidth]{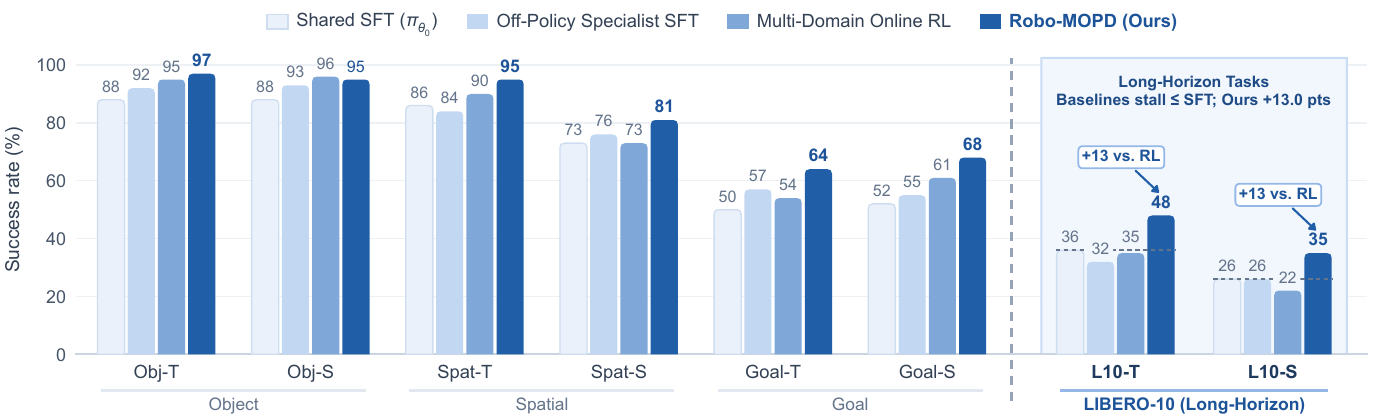}
\vspace{-1mm}
\caption{Success rates (\%) across all eight LIBERO-PRO suites (100 episodes each) under instruction redirection (T) and position swap (S). While Off-Policy Specialist SFT and Multi-Domain Online RL stall or drop below $\pi_{\theta_0}$ on long-horizon LIBERO-10, Robo-MOPD improves over $\pi_{\theta_0}$ (+12 and +9 points on L10-T and L10-S) and leads Online RL by +13 points on both.}
\label{fig:liberopro-suites}
\vspace{-4mm}
\end{figure*}

On LIBERO-PRO (Figure~\ref{fig:main-curves}a), by contrast, where the frozen specialist is substantially stronger at 71.0\%, Robo-MOPD opens a wide margin over Multi-Domain Online RL across rounds (72.9\% vs.\ 65.8\% at Round~8). Multi-Domain Online RL peaks at 67.1\% across all rounds, trails the tabletop specialist by 5.2 points at Round~8, and drops to 28.5\% on LIBERO-10 (below $\pi_{\theta_0}$ at 31.0\% and 11.5 points behind the 40.0\% specialist), consistent with binary episode rewards penalizing valid early subgoals when a later step fails. Using accurate token guidance from the strong tabletop specialist while retaining positive transfer from the kitchen specialist, Robo-MOPD consolidates both harnesses into one 9B checkpoint (halving deployed planner parameter memory relative to two separate 9B specialists) and is the only method that surpasses both specialists simultaneously (+1.9 points on LIBERO-PRO and +2.0 points on RoboCasa365) at 58.9\% Avg.\ (+2.0 points over Domain Specialists and +2.8 points over Multi-Domain Online RL, Figure~\ref{fig:main-curves}c). Gains are largest on Goal (+8.5 points over the specialist, 66.0\% vs.\ 57.5\%), Seen composite kitchen tasks (+7.5 points, 37.5\% vs.\ 30.0\%), and LIBERO-10 (+13.0 points over Multi-Domain Online RL, 41.5\% vs.\ 28.5\%).

\subsection{Analysis}
\label{sec:exp-analysis}

\paragraph{Comparison of consolidation approaches and step efficiency.}
The Consolidation Setup columns in Table~\ref{tab:main-consolidation} compare consolidation strategies varying in on-policy student rollouts and per-token supervision under a 3{,}072-episode Stage~2 budget. While Off-Policy Specialist SFT~\citep{kdrl,openmopd,rldg} raises shorter suites such as Object (88.0\% to 92.5\%) and Goal (51.0\% to 56.0\%) for 52.2\% Avg., supervising only specialist trajectories leaves LIBERO-10 below $\pi_{\theta_0}$ (29.0\% vs.\ 31.0\%) once closed-loop execution drifts into unfamiliar states~\citep{minillm,mopd}. Switching to on-policy student rollouts in Multi-Domain Online RL lifts average accuracy to 56.1\%, yet sparse binary rewards still leave LIBERO-10 at 28.5\%. By combining student rollouts with routed specialist token guidance, Robo-MOPD reaches 41.5\% on LIBERO-10 (+12.5 points over Off-Policy Specialist SFT and +13.0 over Multi-Domain Online RL) and 58.9\% Avg., turning $\pi_{\theta_0}$'s initial 7.5-point deficit behind the specialists into a +2.0-point margin. Dense token supervision also yields higher step efficiency~\citep{mopd,nemotroncascade2}. Group filtering leaves Robo-MOPD with more total updates than Multi-Domain Online RL across eight rounds (268 vs.\ 224 steps). Even so, Robo-MOPD already surpasses eight-round Multi-Domain Online RL (224 steps, 56.1\% Avg.\@) by Round~4 (142 steps, 56.8\% Avg.\@) and Round~5 (173 steps, 58.5\% Avg.\@) with fewer gradient steps (Figure~\ref{fig:suppl-step-efficiency}), and leads on four of the final five rounds despite a temporary dip to 55.1\% at Round~7 (Figure~\ref{fig:main-curves}c).

\paragraph{Task horizon and spatial perturbations.}
Comparing task families in Table~\ref{tab:main-consolidation} and the eight-suite breakdown in Figure~\ref{fig:liberopro-suites} shows that the margin between Robo-MOPD and Multi-Domain Online RL widens on multi-stage and spatially perturbed suites. On short-horizon Object tasks, Multi-Domain Online RL and Robo-MOPD achieve comparable accuracy (95.5\% vs.\ 96.0\%). On tasks requiring sequential subgoals or recovery from displaced layouts, Robo-MOPD outperforms Multi-Domain Online RL by +8.5 points on Goal (66.0\% vs.\ 57.5\%) and +13.0 points on LIBERO-10 (41.5\% vs.\ 28.5\%, including 48.0\% vs.\ 35.0\% under instruction redirection and 35.0\% vs.\ 22.0\% under position swap). Across position swap suites in Spatial, Goal, and LIBERO-10, Robo-MOPD leads Multi-Domain Online RL by 7.0 to 13.0 points, consistent with routed specialist log probabilities supervising intermediate camera checks, pre-grasp alignment, and per-turn tool arguments even when a later subgoal fails.

\paragraph{Failure modes and voluntary termination.}
Robo-MOPD not only fails less often on LIBERO-PRO (27.1\% vs.\ 34.2\% for Multi-Domain Online RL), but it also fails differently (Figure~\ref{fig:suppl-turn-timeout}, with paired trajectories in Table~\ref{tab:suppl-qualitative-trace}). Shared SFT and Online RL almost always fail by running out of turns, time, or context (36.6\% and 31.6\% of all 800 episodes), calling \texttt{finish()} before the goal is met in only 1.0\% and 2.6\% of episodes. Robo-MOPD cuts budget exhaustion to 20.3\% while raising voluntary \texttt{finish()} calls to 6.8\%. Across the 43 archived voluntary-\texttt{finish()} episodes, these are not early give-ups: Robo-MOPD stops late after repeated failed attempts (median turn 78 out of 100), and on those same 43 scenes, Shared SFT, Online RL, and the specialist all fail 62.8\% of the time---Online RL simply runs out of turns on most of them. This shift comes from distillation rather than the prompt. Under the same finish prompt, voluntary \texttt{finish()} stays rare for Shared SFT (1.0\%) and Online RL (2.6\%), but rises steadily across Robo-MOPD rounds (from 1.0\% at Round~0 to 6.8\% at Round~8). Because Robo-MOPD keeps all-fail groups and distills from the specialist---which already calls \texttt{finish()} in 13.4\% of its own failures---the student learns to stop when a hard task cannot be recovered rather than repeating actions until the 100-turn limit.

\begin{figure}[t]
\centering
\includegraphics[width=\columnwidth]{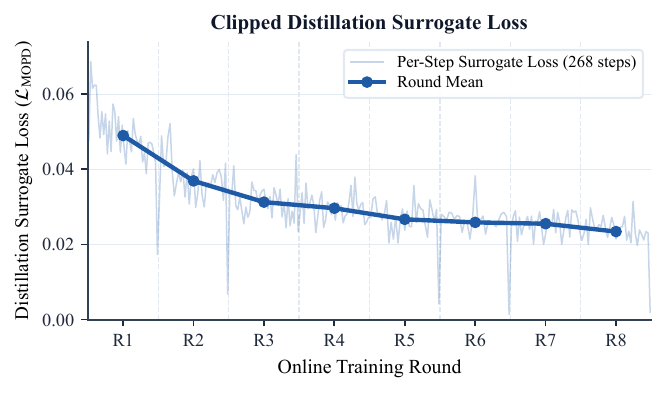}
\vspace{-4mm}
\caption{Clipped distillation surrogate loss $\mathcal{L}_{\mathrm{MOPD}}$ of Robo-MOPD across all 268 gradient steps and per-round means over Rounds~1--8 (R1--R8).}
\label{fig:training-loss}
\vspace{-3mm}
\end{figure}

\begin{figure}[t]
\centering
\includegraphics[width=\columnwidth]{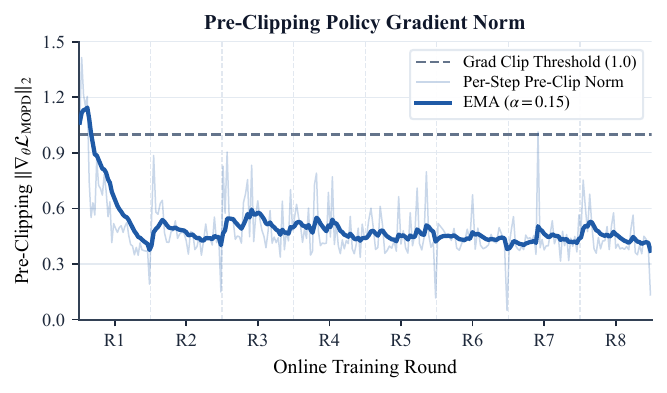}
\vspace{-4mm}
\caption{Pre-clipping gradient $\ell_2$ norm of Robo-MOPD across 268 gradient steps over Rounds~1--8 (R1--R8) with its exponential moving average ($\alpha = 0.15$) relative to the $1.0$ clipping threshold.}
\label{fig:training-gnorm}
\vspace{-3mm}
\end{figure}

\paragraph{Optimization stability.}
Across all 268 update steps, the round-mean MOPD loss decreases monotonically from 0.049 to 0.023, and pre-clipping gradient norms exceed $1.0$ on only 6 steps (Figures~\ref{fig:training-loss}--\ref{fig:training-gnorm}; details in Appendix~\ref{sec:suppl-optim}).

\section{Conclusion}
\label{sec:conclusion}

We studied specialist consolidation for agentic robotics and presented \textbf{Robo-MOPD}, a multi-teacher on-policy distillation framework that unifies frozen specialist planners across distinct robot embodiments, camera layouts, and tool harnesses into one generalist planner operating without external memory or specialist routing. Starting from shared cross-domain SFT initialization, Robo-MOPD runs closed-loop student rollouts and queries the matching specialist for bounded reverse KL token advantages balanced across episode lengths and domains. Across LIBERO-PRO and RoboCasa365, one 9B Robo-MOPD planner reaches 72.9\% and 44.8\% success (58.9\% average), halving parameter memory while surpassing all consolidated baselines in average success, outperforming both 9B specialists, and leading multi-domain online RL by 13.0 points on LIBERO-10. Because Robo-MOPD optimizes specialist token probabilities without task rewards, gains remain bounded when a domain specialist is modest. Future work will explore combining routed token distillation with outcome rewards across uneven teachers, bimanual embodiments, and real-robot deployment.

{
    \small
    \bibliographystyle{ieeenat_fullname}
    \bibliography{main}
}

\clearpage
\makeatletter
\let\Hy@backout\@gobble
\makeatother
\appendix
\twocolumn[\begin{center}\Large\textbf{Appendix}\end{center}\vspace{0.6em}]
\setcounter{equation}{0}
\renewcommand{\theequation}{S\arabic{equation}}
\setcounter{table}{0}
\renewcommand{\thetable}{S\arabic{table}}
\setcounter{figure}{0}
\renewcommand{\thefigure}{S\arabic{figure}}
\renewcommand{\theHequation}{S\arabic{equation}}
\renewcommand{\theHtable}{S\arabic{table}}
\renewcommand{\theHfigure}{S\arabic{figure}}
\renewcommand{\topfraction}{0.95}\renewcommand{\dbltopfraction}{0.95}
\renewcommand{\bottomfraction}{0.7}\renewcommand{\textfraction}{0.05}
\renewcommand{\floatpagefraction}{0.5}\renewcommand{\dblfloatpagefraction}{0.5}
\setcounter{topnumber}{4}\setcounter{dbltopnumber}{4}\setcounter{totalnumber}{6}
\makeatletter 
\setlength{\@fptop}{0pt}\setlength{\@fpsep}{12pt}\setlength{\@fpbot}{0pt plus 1fil}
\setlength{\@dblfptop}{0pt}\setlength{\@dblfpsep}{12pt}\setlength{\@dblfpbot}{0pt plus 1fil}
\makeatother

\providecommand{\robocmark}{\textcolor[rgb]{0.11,0.47,0.22}{\checkmark}}
\providecommand{\roboxmark}{\textcolor[rgb]{0.70,0.09,0.17}{$\times$}}
\providecommand{\robonamark}{\textcolor[rgb]{0.40,0.40,0.40}{--}}

\noindent This appendix is organized as follows.
\begin{itemize}\setlength{\itemsep}{1pt}\setlength{\parskip}{0pt}
\item \textbf{Sec.~\ref{sec:suppl}: Implementation Details and Hyperparameters.} Benchmark splits, held-out scene separation, dataset proportions, and hyperparameters (Table~\ref{tab:suppl-hyperparams}), the two robot harnesses (Table~\ref{tab:suppl-harness-apis}), and Stage~1 loss (Figure~\ref{fig:sft-init-loss}).
\item \textbf{Sec.~\ref{sec:suppl-objectives}: Baseline Objective Formulations.} Objectives for parameter merging, off-policy specialist SFT, and multi-domain online RL.
\item \textbf{Sec.~\ref{sec:suppl-rounds}: Expanded Evaluation Results.} Per-round results (Table~\ref{tab:suppl-all-rounds}), per-suite LIBERO-PRO results (Table~\ref{tab:liberopro-suites}), and step efficiency (Figure~\ref{fig:suppl-step-efficiency}).
\item \textbf{Sec.~\ref{sec:suppl-dynamics}: Failure Modes and Closed-Loop Trajectory Analysis.} Failure breakdown (Figure~\ref{fig:suppl-turn-timeout}) and closed-loop trajectory excerpts (Table~\ref{tab:suppl-qualitative-trace}).
\item \textbf{Sec.~\ref{sec:suppl-optim}: Optimization Dynamics Across Online Rounds.} Distillation loss (Figure~\ref{fig:training-loss}) and gradient norms (Figure~\ref{fig:training-gnorm}) over all 268 update steps.
\end{itemize}

\section{Implementation Details and Hyperparameters}
\label{sec:suppl}

\paragraph{Benchmark task splits and held-out scene separation.}
Following the evaluation protocol of prior robot harness planners~\citep{harnessvla}, LIBERO-PRO~\citep{libero,liberopro} spans 80 tasks organized into eight perturbation suites across Spatial, Object, Goal, and LIBERO-10 under instruction redirection (T) and position swap (S), running 10 tasks on held-out evaluation scene states 1--10 per suite for 800 evaluation episodes. In the tabletop harness, perception queries and analytic motion primitives pair with the \texttt{pi0\_pick} and \texttt{pi0\_doubled} grasp skills, which invoke the frozen $\pi_{0.5}$ VLA policy~\citep{pi05,rlinf}. On RoboCasa365~\citep{robocasa,robocasa365}, we use the mobile kitchen harness and 50-task target split from prior work~\citep{robocasa365,harnessvla} and evaluate held-out seeds~1--5 per task (250 evaluation episodes), comprising 18 atomic tasks (90 episodes), 16 composite tasks seen during RLDX-1 low-level controller pretraining~\citep{rldx} (80 episodes), and 16 composite tasks unseen during RLDX-1 pretraining (80 episodes). Table~\ref{tab:suppl-harness-apis} contrasts the embodiment specifications, camera views, frozen VLA controllers, and callable tool schemas across the two robot harnesses. Here Seen and Unseen denote controller pretraining splits of RLDX-1 rather than task holdouts for the planner, while all Stage~1 and Stage~2 planner rollouts strictly use training scene states $\ge 11$ (states 11--44 on LIBERO-PRO and 11--78 on RoboCasa365 in Stage~1, and states 11--44 in both domains in Stage~2) so evaluation scene states 1--10 and 1--5 are never visited during training. Each frozen 9B domain specialist $\pi_d^{\mathrm{tea}}$ is trained independently from Base Qwen3.5-9B via single-domain supervised fine-tuning followed by six rounds of single-domain online reinforcement learning following prior single-domain post-training~\citep{agenticroborl}.

\begin{figure}[t]
\centering
\includegraphics[width=0.85\columnwidth]{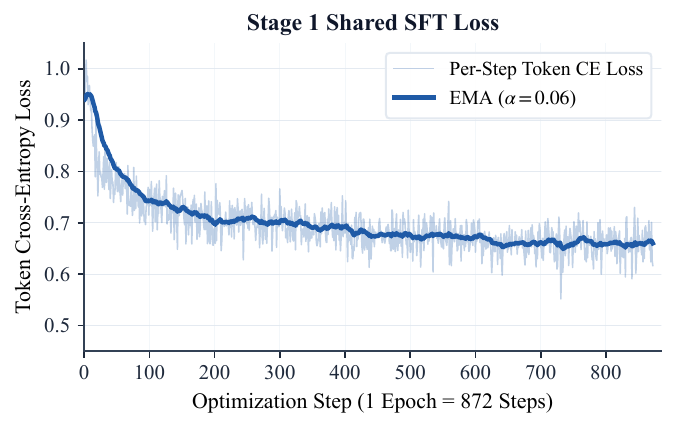}
\caption{Per-step and exponential moving average ($\alpha = 0.06$) token cross-entropy loss during Stage~1 shared initialization ($\pi_{\theta_0}$) over one epoch (872 steps at global batch size 64) on the 55{,}781 verified adapted turns.}
\label{fig:sft-init-loss}
\vspace{-3mm}
\end{figure}

\paragraph{Cross-domain dataset proportions and online domain parity.}
Table~\ref{tab:suppl-hyperparams} summarizes dataset statistics, decoding settings, and shared hyperparameters, and Figure~\ref{fig:sft-init-loss} plots the token cross-entropy loss during Stage~1 initialization ($\pi_{\theta_0}$). In Stage~1, adapting demonstrations collected by Qwen3.8-27B~\citep{qwen38} with task logs~\citep{harnessvla} yields 2{,}450 episodes (40{,}047 verified turns) across LIBERO-PRO and 1{,}350 episodes (15{,}734 verified turns) across RoboCasa365, which we pool directly for one epoch (872 gradient steps at global batch size 64 and learning rate $10^{-5}$) without discarding tabletop tasks or duplicating kitchen trajectories. In Stage~2, all rollout-based methods share a fixed random seed for scene sampling so every method collects the exact same 32 groups of $K = 6$ closed-loop episodes (192 episodes) per domain across Rounds~1--8 (1{,}536 episodes per domain and 3{,}072 episodes in total). During optimization, the normalization $1 / (|\mathcal{D}|\,|\mathcal{E}_d|\,T_e)$ in Eq.~\eqref{eq:mopd-total-loss} divides each retained episode by its selected token count $T_e$~\citep{minillm,mopd,kdrl} so verbose failed rollouts~\citep{turnopd} do not outweigh concise completions, and scales each active set $\mathcal{E}_d$ by $1 / (|\mathcal{D}|\,|\mathcal{E}_d|)$ with $1 / |\mathcal{D}| = 0.5$ to maintain equal preset domain weights~\citep{openmopd}.

\paragraph{Active task curation and clipping bounds.}
From initial screening of $\pi_{\theta_0}$ to exclude saturated and unsolved tasks, we fix a shared Stage~2 active pool of 48 of 80 tasks on LIBERO-PRO and 30 of 50 tasks on RoboCasa365 across all Stage~2 methods. For each retained episode, we uniformly subsample at most $L = 16$ planning turns without replacement in chronological order before performing one pass of minibatch updates at global batch size 128 turns and constant learning rate $10^{-6}$ (with logged per-step losses scaled by a fixed nominal batch constant rather than the turn count of the final minibatch). Across Rounds~1--8, Robo-MOPD retains 222--312 active episodes per round (2{,}244 total active episodes, 33{,}735 turns, and 268 gradient steps), whereas Multi-Domain Online RL retains 210--276 active episodes per round (1{,}896 total active episodes, 28{,}272 turns, and 224 gradient steps). Motivated by heavy-tailed negative log probability gaps on student tokens~\citep{safopd}, we set $[C_{\min}, C_{\max}] = [-3.0, 5.0]$ in Eq.~\eqref{eq:opd-advantage}, applying a tighter negative bound because positive gaps are bounded above when the student already assigns non-trivial probability whereas negative gaps on tokens disfavored by the specialist can spike to large magnitudes and otherwise dominate updates.

\begin{table*}[t]
\centering
\small
\setlength{\tabcolsep}{7.5pt}
\begin{tabular}{@{}ll@{}}
\toprule
Setting / Hyperparameter & Value \\
\midrule
\multicolumn{2}{@{}l}{\textit{Planner backbone, benchmark splits, and evaluation decoding}} \\
Base multimodal backbone & Qwen3.5-9B~\citep{qwen35} \\
Frozen / trainable modules & Vision ViT / LLM + visual merger \\
LIBERO-PRO evaluation split & 80 tasks $\times$ seeds 1--10 (800 episodes, $\pi_{0.5}$ VLA~\citep{pi05}) \\
RoboCasa365 evaluation split & 18 / 16 / 16 tasks $\times$ seeds 1--5 (250 episodes, RLDX-1 VLA~\citep{rldx}) \\
Disjoint training state range & $\ge 11$ (11--44 tabletop and 11--78 kitchen in Stage~1, 11--44 both in Stage~2) \\
Evaluation decoding & Temp.\ $1.0$, $\text{top-}p = 0.95$, $\text{top-}k = 20$, presence pen.\ $1.5$, $M_{\max} = 100$, reset off \\
\midrule
\multicolumn{2}{@{}l}{\textit{Stage~1 cross-domain initialization ($\pi_{\theta_0}$)}} \\
Adapted LIBERO-PRO demonstrations & 2{,}450 episodes (40{,}047 verified turns from Qwen3.8-27B~\citep{qwen38}) \\
Adapted RoboCasa365 demonstrations & 1{,}350 episodes (15{,}734 verified turns from Qwen3.8-27B~\citep{qwen38}) \\
Pooled Stage~1 dataset & 3{,}800 episodes (55{,}781 verified turns) \\
Optimizer and learning rate & AdamW, $1 \times 10^{-5}$ (cosine decay) \\
Global batch size and duration & 64 turns, 1 epoch (872 steps) \\
\midrule
\multicolumn{2}{@{}l}{\textit{Stage~2 shared online and offline rollout training protocol}} \\
Fixed active training task pool & 48 / 80 (LIBERO-PRO), 30 / 50 (RoboCasa365) \\
Online rounds and rollout decoding & 8 rounds (endpoint at Round~8), temp.\ $1.0$, $\text{top-}p = 1.0$, presence pen.\ $0.0$, $M_{\max} = 100$ \\
Rollouts per domain per round & 32 groups $\times$ $K = 6$ (192 episodes, shared sampling seed across methods) \\
Total rollouts across domains & 384 per round (50\% LIBERO-PRO and 50\% RoboCasa365, 3{,}072 total) \\
Domain and episode loss weighting & $1 / |\mathcal{D}| = 0.5$ preset per domain, $1 / T_e$ per retained episode \\
Maximum subsampled turns per episode ($L$) & 16 (uniform, chronological) \\
Token advantage clip $[C_{\min}, C_{\max}]$ & $[-3.0,\, 5.0]$ \\
PPO ratio clip $(\epsilon_-, \epsilon_+)$ & $(0.20,\, 0.28)$ \\
Dual-clip cap on negative advantages ($c$) & $3.0$ \\
Minibatch size and passes & 128 turns, 1 pass \\
Stage~2 learning rate and gradient clip & $1 \times 10^{-6}$ (constant), $1.0$ \\
\bottomrule
\end{tabular}
\caption{Dataset statistics, evaluation decoding settings, benchmark splits, and shared Stage~1 and Stage~2 hyperparameters.}
\label{tab:suppl-hyperparams}
\vspace{-3mm}
\end{table*}

\begin{table*}[t]
\centering
\footnotesize
\setlength{\tabcolsep}{5.5pt}
\renewcommand{\arraystretch}{1.00}
\begin{tabular}{@{}p{0.152\textwidth}p{0.402\textwidth}p{0.402\textwidth}@{}}
\toprule
\textbf{Dimension} &
\textbf{LIBERO-PRO Tabletop Harness} &
\textbf{RoboCasa365 Mobile Kitchen Harness} \\
\midrule
\multicolumn{3}{@{}l}{\textit{Embodiment, visual sensing, frozen low-level VLA, and evaluation protocol}} \\
\raggedright Benchmark \& split &
\raggedright 80 tasks across 8 perturbation suites (Spatial, Object, Goal, LIBERO-10 $\times$ T/S), held-out seeds 1--10 (800 episodes) &
\raggedright 50 target-split kitchen tasks (18 Atomic, 16 Composite Seen, 16 Composite Unseen), held-out seeds 1--5 (250 episodes) \tabularnewline
\addlinespace[1.0pt]
\raggedright Robot \& action space &
\raggedright Fixed-base 7-DoF Franka Panda arm with parallel-jaw gripper, 7-D \texttt{OSC\_POSE} delta action $[\Delta\mathbf{p}_3, \Delta\mathbf{r}_3, g_1]$ ($g{=}{+}1$ close, ${-}1$ open) &
\raggedright \texttt{PandaOmron} mobile manipulator (7-DoF arm, gripper, wheeled planar base, torso), 12-D action $[\Delta\mathbf{p}_3, \Delta\mathbf{r}_3, g_1, \mathbf{b}_4, m_1]$ (\texttt{gripper="hold"} servo) \tabularnewline
\addlinespace[1.0pt]
\raggedright Cameras \& resolution &
\raggedright Two RGB-D views (\texttt{agentview} and \texttt{wrist}), $256{\times}256$ state RGB plus on-demand $1024{\times}1024$ RGB-D and metric 3D world maps &
\raggedright Three RGB-D views (\texttt{agentview}, \texttt{navview}, and \texttt{wrist}), $256{\times}256$ RGB plus per-step metric 3D world maps \tabularnewline
\addlinespace[1.0pt]
\raggedright Frozen VLA controller &
\raggedright Frozen $\pi_{0.5}$~\citep{pi05,rlinf} predicting continuous action chunks conditioned on tool-supplied sub-instruction \texttt{prompt} &
\raggedright Frozen RLDX-1~\citep{rldx} consuming 4-frame 3-camera RGB history ($256{\times}256$) and relative base/EEF state (8 steps per chunk) \tabularnewline
\addlinespace[1.0pt]
\raggedright Episode protocol &
\raggedright Single-attempt evaluation with turn cap $M_{\max}{=}100$, \texttt{reset} disabled, and no access to prior task logs &
\raggedright Single-attempt evaluation with turn cap $M_{\max}{=}100$, \texttt{reset} disabled, and no access to prior task logs \tabularnewline
\midrule
\multicolumn{3}{@{}l}{\textit{Callable domain perception, motion, navigation, and VLA skill tools (required arguments marked with \texttt{*})}} \\
\raggedright Visual \& 3D spatial perception queries &
\raggedright
\textbf{(1)}~\texttt{view\_env\_state}(\texttt{step}) returns EEF pose, finger gap, and RGB-D views.
\textbf{(2)}~\texttt{view\_camera\_meta}(\texttt{camera}, \texttt{step}) returns intrinsics $\mathbf{K}$ and extrinsics $\mathbf{T}_{\mathrm{c2w}}$.
\textbf{(3)}~\texttt{segment}(\texttt{prompt}, \texttt{camera}, \texttt{step}, \texttt{point}, \texttt{min\_score}) returns SAM3~\citep{sam3} mask and 3D world centroid.
\textbf{(4)}~\texttt{back\_project}(\texttt{row}, \texttt{col}, \texttt{step}, \texttt{camera}, \texttt{resolution}, \texttt{row\_range}, \texttt{col\_range}, \texttt{z\_min}, \texttt{z\_max}) projects pixels or $z$-filtered regions to 3D. &
\raggedright
\textbf{(1)}~\texttt{view\_env\_state}(\texttt{step}) returns EEF/base pose, task predicates, and 3-camera RGB/world maps.
\textbf{(2)}~\texttt{view\_camera\_meta}(\texttt{camera}, \texttt{step}) returns camera $\mathbf{K}$ and $\mathbf{T}_{\mathrm{c2w}}$.
\textbf{(3)}~\texttt{back\_project}(\texttt{row*}, \texttt{col*}, \texttt{step}, \texttt{camera}, \texttt{resolution}) queries single-pixel 3D coordinates.
\textbf{(4)}~\texttt{back\_project\_batch}(\texttt{pixels*}, \texttt{step}, \texttt{camera}, \texttt{resolution}) runs multi-pixel 3D lookup.
\textbf{(5)}~\texttt{query\_world\_map}(\texttt{z\_min}, \texttt{z\_max}, \texttt{x\_range}, \texttt{y\_range}, \texttt{camera}, \texttt{resolution}, \texttt{min\_cluster\_size}) finds 3D connected-component clusters. \tabularnewline
\addlinespace[1.2pt]
\raggedright Analytic arm \& gripper motion primitives &
\raggedright
\textbf{(1)}~\texttt{move\_to}(\texttt{xyz*}, \texttt{gripper}, \texttt{tol}, \texttt{step\_clip}, \texttt{max\_steps}, \texttt{action\_scale}, \texttt{target\_yaw}, \texttt{yaw\_step\_clip}) runs world-XYZ OSC servo.
\textbf{(2)}~\texttt{move\_pose}(\texttt{xyz*}, \texttt{target\_pitch}, \texttt{target\_yaw}, \texttt{gripper}, \texttt{step\_clip}, \texttt{pitch\_step}, \texttt{yaw\_step}, \texttt{tol}, \texttt{ori\_tol}, \texttt{action\_scale}, \texttt{max\_steps}) runs coupled 5-DoF servo.
\textbf{(3)}~\texttt{rotate\_wrist}(\texttt{target\_yaw}, \texttt{delta\_yaw}, \texttt{gripper}, \texttt{max\_steps}, \texttt{tol}, \texttt{step\_clip}) rotates world-$z$ yaw.
\textbf{(4)}~\texttt{rotate\_pitch}(\texttt{target\_pitch}, \texttt{delta\_pitch}, \texttt{gripper}, \texttt{max\_steps}, \texttt{tol}, \texttt{step\_clip}) tilts world-$x$ pitch.
\textbf{(5)}~\texttt{set\_gripper}(\texttt{gripper}, \texttt{steps}) commands fingers (${+}1$ / ${-}1$).
\textbf{(6)}~\texttt{release}(\texttt{max\_steps}) opens fingers in place. &
\raggedright
\textbf{(1)}~\texttt{move\_to}(\texttt{xyz*}, \texttt{gripper}, \texttt{step\_clip}, \texttt{max\_steps}, \texttt{tol}) runs online-Jacobian world-XYZ servo (\texttt{gripper="hold"} default).
\textbf{(2)}~\texttt{move\_delta}(\texttt{dxyz*}, \texttt{gripper}, \texttt{step\_clip}, \texttt{max\_steps}) executes relative world EEF displacement.
\textbf{(3)}~\texttt{rotate\_pitch}(\texttt{target\_pitch}, \texttt{gripper}, \texttt{n}) tilts wrist pitch over $n$ steps.
\textbf{(4)}~\texttt{set\_gripper}(\texttt{gripper}, \texttt{steps}) closes or opens fingers.
\textbf{(5)}~\texttt{release}(\texttt{steps}) opens gripper (\texttt{gripper=-1}) in place.
\textbf{(6)}~\texttt{scripted\_grasp}(\texttt{xyz*}, \texttt{approach\_z}, \texttt{grasp\_z\_offset}, \texttt{step\_clip}) runs 4-stage analytic open-and-hover, descend, close, and lift grasp. \tabularnewline
\addlinespace[1.2pt]
\raggedright Mobile base navigation &
\raggedright
\emph{None} (fixed tabletop base, workspace spanned directly by arm kinematics). &
\raggedright
\textbf{(1)}~\texttt{navigate\_to}(\texttt{xy*}, \texttt{tol}, \texttt{max\_steps}, \texttt{gripper}) drives planar base to world $(x, y)$ with grip hold.
\textbf{(2)}~\texttt{move\_base}(\texttt{forward}, \texttt{lateral}, \texttt{turn}, \texttt{steps}, \texttt{gripper}) sends robot-local planar velocity $(v_x, v_y, \omega_z)$. \tabularnewline
\addlinespace[1.2pt]
\raggedright Frozen VLA closed-loop skill wrappers &
\raggedright
\textbf{(1)}~\texttt{pi0\_pick}(\texttt{prompt*}, \texttt{max\_chunks}, \texttt{lift\_thresh}, \texttt{gripper\_closed\_thresh}) runs closed-loop $\pi_{0.5}$ grasp and exits on verified lift and closed finger gap.
\textbf{(2)}~\texttt{pi0\_doubled}(\texttt{prompt*}, \texttt{max\_chunks}) runs closed-loop $\pi_{0.5}$ contact skill until task predicate holds. &
\raggedright
\textbf{(1)}~\texttt{rldx\_skill}(\texttt{prompt*}, \texttt{base\_clip}, \texttt{max\_chunks}, \texttt{use\_prompt}, \texttt{force\_reset}) runs unclamped whole-body RLDX-1 rollout (arm plus base, capped at 40 chunks in deployment).
\textbf{(2)}~\texttt{rldx\_arm}(\texttt{prompt*}, \texttt{base\_clip=0.1}, \texttt{max\_chunks}) clamps planar base ($|\mathbf{b}| \le 0.1$) for near-contact alignment. \tabularnewline
\addlinespace[1.2pt]
\raggedright Episode termination &
\raggedright
\texttt{finish}(\texttt{status*}, \texttt{summary*}) terminates the episode and triggers simulator goal check. &
\raggedright
\texttt{finish}(\texttt{status*}, \texttt{summary*}) terminates the episode and triggers simulator goal check. \tabularnewline
\bottomrule
\end{tabular}
\caption{Comparison of the two distinct robot harnesses operated by a single 9B Robo-MOPD planner without external specialist routing. At each planning turn $m$, the active harness injects its domain-native JSON tool schema into prompt context $x_{e,m}$, while exploration-only \texttt{reset} remains disabled across all evaluations.}
\label{tab:suppl-harness-apis}
\vspace{-3mm}
\end{table*}

\section{Baseline Objective Formulations}
\label{sec:suppl-objectives}

\paragraph{Specialist merging in parameter space.}
Because both frozen 9B specialists $\{\theta_d^{\mathrm{tea}}\}_{d \in \mathcal{D}}$ are post-trained from the same pretrained checkpoint $\theta_{\mathrm{base}}$ (Base Qwen3.5-9B), training-free consolidation combines their task vectors in parameter space~\citep{modelsoups,taskarithmetic},
\begin{equation}
\theta_{\mathrm{merge}}(\alpha)
=
\theta_{\mathrm{base}}
+
\alpha \sum_{d \in \mathcal{D}}
\bigl(\theta_d^{\mathrm{tea}} - \theta_{\mathrm{base}}\bigr),
\label{eq:task-arithmetic}
\end{equation}
producing uniform \textbf{Weight Averaging}~\citep{modelsoups} at $\alpha = 1 / |\mathcal{D}| = 0.5$ and unscaled \textbf{Task Arithmetic}~\citep{taskarithmetic} at $\alpha = 1.0$ (note that for $|\mathcal{D}| = 2$ specialists, tuning $\alpha$ to $0.5$ in Task Arithmetic is algebraically identical to Weight Averaging).

\paragraph{Off-policy specialist SFT.}
Following rejection sampling fine-tuning on teacher rollouts~\citep{kdrl,openmopd} and specialist rollout distillation in~\citet{rldg}, \textbf{Off-Policy Specialist SFT}~\citep{mopd,kdrl,openmopd,rldg} collects closed-loop trajectories $\tau_e \sim \pi_d^{\mathrm{tea}}$ directly from each frozen 9B specialist on the exact 3{,}072-episode Stage~2 training scene schedule over states 11--44 without prior task logs. Excluding 10 rollouts that timed out in simulation leaves 3{,}062 valid episodes across 1{,}530 LIBERO-PRO and 1{,}532 RoboCasa365 runs, from which failed rollouts ($R_e = 0$) are discarded so forward cross-entropy does not imitate unsuccessful actions. The resulting successful sets $\mathcal{E}_d^{\mathrm{off}}$ retain 1{,}103 LIBERO-PRO and 844 RoboCasa365 episodes (1{,}947 episodes in total), and uniformly subsampling at most $L = 16$ turn indices in chronological order per episode to form $\mathcal{M}_e$ yields 23{,}172 selected planning turns. Applying the same domain and episode token normalization as Eq.~\eqref{eq:mopd-total-loss} with equal $0.5$ preset weight per domain, we update $\pi_{\theta_0}$ for one pass (182 gradient steps at global batch size 128 and constant learning rate $10^{-6}$) and evaluate the final checkpoint after the pass~\citep{flowvlamopd}, minimizing forward token cross-entropy,
\begin{equation}
\begin{aligned}
\mathcal{L}_{\mathrm{Off\text{-}SFT}}(\theta)
&= -\frac{1}{|\mathcal{D}|}
\sum_{d \in \mathcal{D}}
\frac{1}{|\mathcal{E}_d^{\mathrm{off}}|}
\sum_{e \in \mathcal{E}_d^{\mathrm{off}}} \\
&\quad
\frac{1}{T_e}
\sum_{m \in \mathcal{M}_e}
\sum_{t=1}^{n_{e,m}}
\log \pi_\theta(y_{e,m,t} \mid x_{e,m}, y_{e,m,<t}).
\end{aligned}
\label{eq:offpolicy-sft}
\end{equation}

\paragraph{Multi-domain online reinforcement learning.}
Under the identical eight-round student rollout schedule from $\pi_{\theta_0}$, \textbf{Multi-Domain Online RL} replaces $A_{e,m,t}^{\mathrm{opd}}$ with the group leave-one-out episode advantage~\citep{rloo,agenticroborl} computed from binary completion indicators $R_e \in \{0, 1\}$ across the $K$ attempts in rollout group $\mathcal{G}$,
\begin{equation}
A_e^{\mathrm{task}}
=
R_e
-
\frac{1}{K - 1}
\sum_{e' \in \mathcal{G} \setminus \{e\}} R_{e'}.
\label{eq:rl-advantage}
\end{equation}
Because both fully solved groups ($\sum_{e \in \mathcal{G}} R_e = K$) and groups with zero successes ($\sum_{e \in \mathcal{G}} R_e = 0$) yield $A_e^{\mathrm{task}} = 0$ for every episode, only mixed-outcome groups ($0 < \sum_{e \in \mathcal{G}} R_e < K$) form the active episode set $\mathcal{E}_d^{\mathrm{RL}}$~\citep{dapo}. Substituting $A_e^{\mathrm{task}}$ for $A_{e,m,t}^{\mathrm{opd}}$ across all selected response tokens of episode $e$ inside the clipped surrogate $\ell_{e,m,t}^{\mathrm{task}}(\theta)$ of Eq.~\eqref{eq:token-ppo-loss} under the same domain and episode token normalization yields
\begin{equation}
\begin{aligned}
\mathcal{L}_{\mathrm{RL}}(\theta)
&= \frac{1}{|\mathcal{D}|}
\sum_{d \in \mathcal{D}}
\frac{1}{|\mathcal{E}_d^{\mathrm{RL}}|}
\sum_{e \in \mathcal{E}_d^{\mathrm{RL}}} \\
&\quad
\frac{1}{T_e}
\sum_{m \in \mathcal{M}_e}
\sum_{t=1}^{n_{e,m}}
\ell_{e,m,t}^{\mathrm{task}}(\theta).
\end{aligned}
\label{eq:rl-loss}
\end{equation}

\section{Expanded Evaluation Results}
\label{sec:suppl-rounds}

\begin{figure}[t]
\centering
\includegraphics[width=0.8\columnwidth]{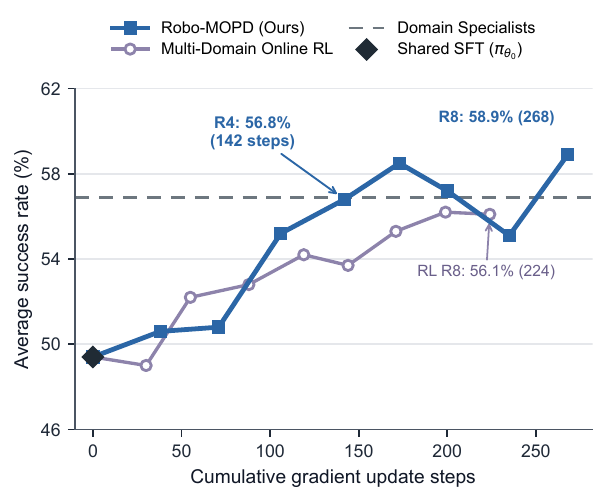}
\caption{Two-domain average evaluation success rate (\%) plotted against cumulative Stage~2 gradient update steps across Rounds~1--8. Robo-MOPD at Round~4 (142 steps, 56.8\% Avg.\@) already outperforms eight-round Multi-Domain Online RL (224 steps, 56.1\% Avg.\@).}
\label{fig:suppl-step-efficiency}
\vspace{-3mm}
\end{figure}

\paragraph{Per-round results, suite breakdown, and step efficiency.}
Table~\ref{tab:suppl-all-rounds} reports complete checkpoint evaluation results from Shared SFT $\pi_{\theta_0}$ (Round~0) through Round~8 across LIBERO-PRO (800 episodes) and RoboCasa365 (250 episodes), corresponding to Figure~\ref{fig:main-curves} in the main text, and Table~\ref{tab:liberopro-suites} breaks down Round~8 accuracy across all eight LIBERO-PRO perturbation suites. In Figure~\ref{fig:suppl-step-efficiency}, we also plot two-domain average success against cumulative Stage~2 gradient update steps~\citep{mopd,vlaopd} because group filtering leaves Robo-MOPD with more total updates than Multi-Domain Online RL across eight rounds (268 vs.\ 224 steps). Robo-MOPD already exceeds full eight-round Multi-Domain Online RL (224 steps, 56.1\% Avg.\@) at Round~4 (142 steps, 56.8\% Avg.\@), showing that its higher final accuracy does not come from taking more update steps.

\begin{table*}[t]
\centering
\small
\setlength{\tabcolsep}{7.6pt}
\begin{tabular}{@{}c cc cc cc@{}}
\toprule
& \multicolumn{2}{c}{LIBERO-PRO (800 eps, \%)} & \multicolumn{2}{c}{RoboCasa365 (250 eps, \%)} & \multicolumn{2}{c}{Two-Domain Avg.\ (\%)} \\
\cmidrule(lr){2-3} \cmidrule(lr){4-5} \cmidrule(lr){6-7}
Round & Online RL & Robo-MOPD & Online RL & Robo-MOPD & Online RL & Robo-MOPD \\
\midrule
0 ($\pi_{\theta_0}$) & 62.4 & 62.4 & 36.4 & 36.4 & 49.4 & 49.4 \\
1 & 62.8 & 64.4 & 35.2 & 36.8 & 49.0 & 50.6 \\
2 & 63.9 & 64.0 & 40.4 & 37.6 & 52.2 & 50.8 \\
3 & 66.4 & 69.5 & 39.2 & 40.8 & 52.8 & 55.2 \\
4 & 63.9 & 69.1 & 44.4 & 44.4 & 54.2 & 56.8 \\
5 & 61.8 & \textbf{73.8} & 45.6 & 43.2 & 53.7 & 58.5 \\
6 & 66.6 & 73.6 & 44.0 & 40.8 & 55.3 & 57.2 \\
7 & 67.1 & 68.9 & 45.2 & 41.2 & 56.2 & 55.1 \\
8 & 65.8 & 72.9 & \textbf{46.4} & 44.8 & 56.1 & \textbf{58.9} \\
\bottomrule
\end{tabular}
\caption{Complete checkpoint evaluation success rates (\%) from Shared SFT $\pi_{\theta_0}$ (Round~0) through Round~8 for Multi-Domain Online RL (Online RL) and Robo-MOPD across LIBERO-PRO (800 episodes) and RoboCasa365 (250 episodes), where Two-Domain Avg.\ is the mean of the reported LIBERO-PRO and RoboCasa365 columns. All endpoint comparisons in the main text use Round~8, and bold marks the highest score across all rounds and methods in each block.}
\label{tab:suppl-all-rounds}
\vspace{-3mm}
\end{table*}

\begin{table*}[t]
\centering
\small
\setlength{\tabcolsep}{5.2pt}
\begin{tabular}{@{}lccccccccc@{}}
\toprule
Method & Spat-T & Spat-S & Obj-T & Obj-S & Goal-T & Goal-S & L10-T & L10-S & Overall \\
\midrule
\multicolumn{10}{@{}l}{\textit{Reference policies}} \\
Base Qwen3.5-9B & 22.0 & 17.0 & 19.0 & 15.0 & 24.0 & 35.0 & 9.0 & 4.0 & 18.1 \\
Shared SFT ($\pi_{\theta_0}$) & 86.0 & 73.0 & 88.0 & 88.0 & 50.0 & 52.0 & 36.0 & 26.0 & 62.4 \\
LIBERO-PRO Specialist & 96.0 & 86.0 & 96.0 & 95.0 & 54.0 & 61.0 & 41.0 & 39.0 & 71.0 \\
\midrule
\multicolumn{10}{@{}l}{\textit{Consolidated generalist 9B planners}} \\
Weight Averaging ($\alpha = 0.5$) & 83.0 & 77.0 & 76.0 & 86.0 & 51.0 & 53.0 & 27.0 & 15.0 & 58.5 \\
Task Arithmetic ($\alpha = 1.0$) & 22.0 & 28.0 & 45.0 & 51.0 & 22.0 & 25.0 & 5.0 & 3.0 & 25.1 \\
Off-Policy Specialist SFT & 84.0 & 76.0 & 92.0 & 93.0 & 57.0 & 55.0 & 32.0 & 26.0 & 64.4 \\
Multi-Domain Online RL & 90.0 & 73.0 & 95.0 & \textbf{96.0} & 54.0 & 61.0 & 35.0 & 22.0 & 65.8 \\
\textbf{Robo-MOPD (Ours)} & \textbf{95.0} & \textbf{81.0} & \textbf{97.0} & 95.0 & \textbf{64.0} & \textbf{68.0} & \textbf{48.0} & \textbf{35.0} & \textbf{72.9} \\
\bottomrule
\end{tabular}
\caption{Per-suite success rates (\%) on LIBERO-PRO for reference policies and consolidated planners (Round~8 for online methods), with 100 evaluation episodes per suite across Spatial (Spat), Object (Obj), Goal, and LIBERO-10 (L10) under instruction redirection (T) and position swap (S). Bold marks the best consolidated 9B planner in each column.}
\label{tab:liberopro-suites}
\vspace{-3mm}
\end{table*}

\begin{figure}[t]
\centering
\includegraphics[width=0.8\columnwidth]{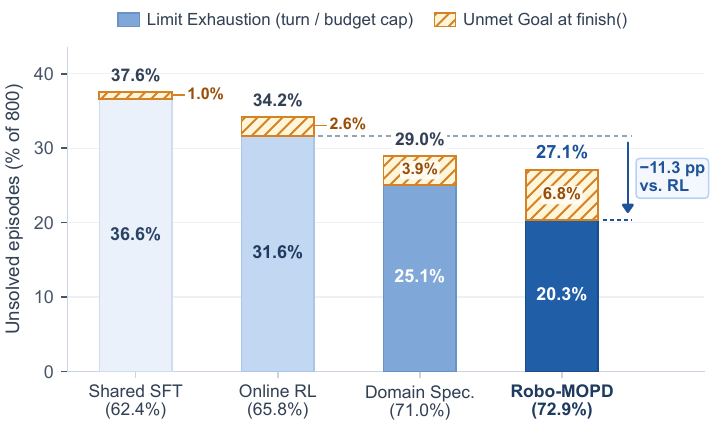}
\caption{Breakdown of unsolved episodes (\% of 800) on LIBERO-PRO into limit exhaustion (reaching the 100-turn, wall-clock, or context cap without calling \texttt{finish()}) and voluntary \texttt{finish()} calls with an unmet task goal across Shared SFT ($\pi_{\theta_0}$), Round~8 Multi-Domain Online RL, LIBERO-PRO Specialist, and Round~8 Robo-MOPD, with failure shares computed from archived episode logs.}
\label{fig:suppl-turn-timeout}
\vspace{-3mm}
\end{figure}

\section{Failure Modes and Closed-Loop Trajectory Analysis}
\label{sec:suppl-dynamics}

\paragraph{Failure mode breakdown on LIBERO-PRO.}
Figure~\ref{fig:suppl-turn-timeout} decomposes unsolved episodes on LIBERO-PRO (800 episodes) into limit exhaustion (reaching the 100-turn, wall-clock, or context limit without calling \texttt{finish()}) and voluntary \texttt{finish()} calls with an unmet task goal~\citep{captain}. Nearly all failures in Shared SFT (36.6\% of 37.6\%) and Round~8 Multi-Domain Online RL (31.6\% of 34.2\%) come from repeating actions until exhausting execution limits. Distilling from the specialist lowers limit exhaustion in Round~8 Robo-MOPD to 20.3\% ($-$11.3 pp vs.\ Online RL, and below 25.1\% for the specialist), with the remaining 6.8\% calling \texttt{finish()} before the simulator goal predicate is satisfied (vs.\ 3.9\% for the specialist) for the lowest overall failure rate of 27.1\%. Table~\ref{tab:suppl-qualitative-trace} illustrates a closed-loop trajectory excerpt and paired scenes across both harnesses where Online RL hits the 100-turn cap while Robo-MOPD completes the task in 6--27 turns.

\begin{table*}[t]
\centering
\footnotesize
\setlength{\tabcolsep}{4.8pt}
\renewcommand{\arraystretch}{1.08}
\begin{tabular}{@{}p{0.078\textwidth}p{0.286\textwidth}p{0.302\textwidth}p{0.272\textwidth}@{}}
\toprule
\multicolumn{4}{@{}p{\textwidth}@{}}{\textbf{(a) Mid-episode grounding, error recovery, and grasp verification excerpt (Turns 3--11) of Robo-MOPD on LIBERO-PRO (\texttt{libero\_object\_task}, \texttt{task=5}, \texttt{seed=7})} \emph{``Pick the bbq sauce and place it in the basket''}} \\
\midrule
\textbf{Turns} &
\textbf{Planner Grounding \& Reasoning} &
\textbf{Invoked Harness Tool \& Arguments} &
\textbf{Simulator Observation \& Return} \\
\midrule
\raggedright T3 \newline (Step 0) &
\raggedright Grounds target bottle in $1024{\times}1024$ \texttt{agentview}, separating brown bottle from butter and pudding. &
\raggedright \texttt{segment}(\texttt{prompt="the brown BBQ sauce bottle"}, \texttt{camera="agentview"}, \texttt{step=0}) &
\raggedright \texttt{found=true}, \texttt{score=0.84}, \texttt{world\_xyz} $[-0.1912, -0.0713, 0.0530]$ verified on bottle body. \tabularnewline
\addlinespace[1.5pt]
\raggedright T4--T6 \newline (Step 0) &
\raggedright Queries basket window $[460,580]{\times}[760,900]$, detects wall bias, and triangulates two interior cloth pixels. &
\raggedright \texttt{back\_project}(\texttt{row\_range=[460,580]}, \texttt{col\_range=[760,900]}, \texttt{z\_min=0.01}, \texttt{z\_max=0.12}), \newline
\texttt{back\_project}(\texttt{row=490}, \texttt{col=820}) &
\raggedright Region \texttt{median\_xyz} $[-0.0393, 0.2678, 0.0946]$ (\texttt{n\_valid=12883}), pixel $(490, 820) \to [-0.0417, 0.2725, 0.0884]$. \tabularnewline
\addlinespace[1.5pt]
\raggedright T7 \newline (Step 0) &
\raggedright Probes floor filter \texttt{z\_max=0.06}, infers from 0-pixel return that cloth liner sits at $z{\approx}0.06\text{--}0.09\text{ m}$, and locks cavity $(-0.045, 0.28)$. &
\raggedright \texttt{back\_project}(\texttt{row\_range=[490,550]}, \texttt{col\_range=[800,880]}, \texttt{step=0}, \texttt{z\_max=0.06}) &
\raggedright Returns \emph{``too few valid pixels in region after z-filter (0)''} (\texttt{n\_valid\_before\_zfilter=4800}), confirming liner height $>0.06\text{ m}$. \tabularnewline
\addlinespace[1.5pt]
\raggedright T8--T9 \newline (Steps 1--2) &
\raggedright Servos open gripper to hover over $(-0.19, -0.071, 0.22)$, checks bottle in \texttt{wrist} view, and runs capped $\pi_{0.5}$ grasp. &
\raggedright \texttt{move\_to}(\texttt{xyz=[-0.19,-0.071,0.22]}, \texttt{gripper=-1}, \texttt{tol=0.015}, \texttt{step\_clip=0.02}), \newline
\texttt{pi0\_pick}(\texttt{prompt="pick up the bbq sauce bottle"}, \texttt{max\_chunks=12}) &
\raggedright Hover reached in 13 steps, \texttt{pi0\_pick} returns \texttt{success=true}, \texttt{chunks\_used=8}, \texttt{peak\_lift\_m=0.0683}, \texttt{min\_gripper\_opening=0.0364}. \tabularnewline
\addlinespace[1.5pt]
\raggedright T10--T11 \newline (Steps 3--4) &
\raggedright Locks grip via \texttt{set\_gripper(+1)} after modest $6.8\text{ cm}$ lift, then lifts vertically to $z{=}0.28\text{ m}$ to verify clearance. &
\raggedright \texttt{set\_gripper}(\texttt{gripper=1}, \texttt{steps=10}), \newline
\texttt{move\_to}(\texttt{xyz=[-0.2,-0.056,0.28]}, \texttt{gripper=1}, \texttt{tol=0.012}, \texttt{step\_clip=0.015}) &
\raggedright Finger qpos tightens to $[0.0200, -0.0160]$, EEF reaches $z{=}0.2685\text{ m}$, \texttt{wrist} view confirms bottle held above empty table. \tabularnewline
\bottomrule
\end{tabular}

\vspace{3pt}

\begin{tabular}{@{}p{0.325\textwidth}p{0.184\textwidth}p{0.238\textwidth}p{0.191\textwidth}@{}}
\toprule
\multicolumn{4}{@{}p{\textwidth}@{}}{\textbf{(b) Paired closed-loop episode outcomes on identical held-out \texttt{(task, seed)} scenes across consolidated 9B planners}} \\
\midrule
\textbf{Domain \& Held-Out Scene (\texttt{task, seed})} &
\textbf{Off-Policy Spec.\ SFT} &
\textbf{Multi-Domain Online RL (R8)} &
\textbf{Robo-MOPD (R8, Ours)} \\
\midrule
\raggedright LIBERO-PRO \texttt{libero\_object\_task} (\texttt{t=5, s=7}) &
Solved ($1{,}097.0\text{ s}$) &
Solved ($53\text{ turns}$, $501.1\text{ s}$) &
\textbf{Solved ($\mathbf{33}\text{ turns}$, $\mathbf{299.0}\text{ s}$)} \tabularnewline
\raggedright LIBERO-PRO \texttt{libero\_goal\_task} (\texttt{t=8, s=2}) &
Failed (\texttt{solved=false}) &
Failed ($100\text{ turns}$, cap, $1{,}455.4\text{ s}$) &
\textbf{Solved ($\mathbf{17}\text{ turns}$, $\mathbf{80.3}\text{ s}$)} \tabularnewline
\raggedright LIBERO-PRO \texttt{libero\_goal\_swap} (\texttt{t=4, s=7}) &
Failed (\texttt{solved=false}) &
Failed ($100\text{ turns}$, cap, $1{,}984.5\text{ s}$) &
\textbf{Solved ($\mathbf{27}\text{ turns}$, $\mathbf{179.5}\text{ s}$)} \tabularnewline
\raggedright LIBERO-PRO \texttt{libero\_10\_swap} (\texttt{t=2, s=6}) &
Failed (\texttt{solved=false}) &
Failed ($100\text{ turns}$, cap, $1{,}354.8\text{ s}$) &
\textbf{Solved ($\mathbf{22}\text{ turns}$, $\mathbf{201.5}\text{ s}$)} \tabularnewline
\raggedright LIBERO-PRO \texttt{libero\_10\_task} (\texttt{t=2, s=4}) &
Failed (\texttt{solved=false}) &
Failed ($100\text{ turns}$, cap, $1{,}737.5\text{ s}$) &
\textbf{Solved ($\mathbf{23}\text{ turns}$, $\mathbf{239.9}\text{ s}$)} \tabularnewline
\raggedright RoboCasa365 Seen \texttt{WashLettuce} (\texttt{seed=2}) &
Failed (\texttt{solved=false}) &
Failed ($100\text{ turns}$, cap, $5{,}512.6\text{ s}$) &
\textbf{Solved ($\mathbf{8}\text{ turns}$, $\mathbf{251.4}\text{ s}$)} \tabularnewline
\raggedright RoboCasa365 Seen \texttt{RinseSinkBasin} (\texttt{seed=3}) &
Failed (\texttt{solved=false}) &
Failed ($100\text{ turns}$, cap, $3{,}067.5\text{ s}$) &
\textbf{Solved ($\mathbf{6}\text{ turns}$, $\mathbf{116.3}\text{ s}$)} \tabularnewline
\bottomrule
\end{tabular}
\caption{Closed-loop multi-turn trajectory excerpt and paired scene-level outcome comparison across consolidated 9B planners. \textbf{(a)}~Mid-episode planning excerpt (Turns~3--11 of the 33-turn trajectory) of Robo-MOPD on held-out \texttt{libero\_object\_task} (\texttt{task=5}, \texttt{seed=7}), illustrating SAM3~\citep{sam3} segmentation, multi-query 3D spatial grounding, recovery from an empty $z$-filtered \texttt{back\_project} query (Turn~7), and $\pi_{0.5}$ grasp verification via finger-gap and \texttt{wrist}-camera feedback (Turns~9--11) prior to bounded basket placement. \textbf{(b)}~Paired outcomes on identical held-out \texttt{(task, seed)} episodes across tabletop and mobile kitchen suites: Robo-MOPD solves the single-object scene in fewer turns (33 vs.\ 53) and completes all six multi-stage scenes concisely (6--27 turns), where Off-Policy Specialist SFT fails under state drift and Multi-Domain Online RL exhausts the $M_{\max}{=}100$ turn cap.}
\label{tab:suppl-qualitative-trace}
\vspace{-3mm}
\end{table*}

\section{Optimization Dynamics Across Online Rounds}
\label{sec:suppl-optim}

Figures~\ref{fig:training-loss}--\ref{fig:training-gnorm} in the main paper track optimization dynamics across all 268 update steps over Rounds~1--8~\citep{nemotroncascade2,rethinkingopd}. Even as the student visits newly sampled simulator states each round~\citep{flowvlamopd}, the round-mean clipped surrogate loss $\mathcal{L}_{\mathrm{MOPD}}$ drops monotonically from 0.049 to 0.031, 0.027, and 0.023 at Rounds~1, 3, 5, and 8 (Figure~\ref{fig:training-loss}). Meanwhile, the round-mean absolute token log probability gap before PPO ratio clipping~\citep{openmopd} contracts monotonically from 0.148 to 0.094 with round-mean PPO token clipping rates of 0.2\%--3.0\%. Pre-clipping gradient $\ell_2$ norms peak at 1.41 in early Round~1 (round mean 0.61), exceed the $1.0$ threshold on only 6 of 268 steps (five in Round~1 and one at 1.01 in Round~7), and maintain stable round means of 0.43--0.53 across Rounds~2--8 (Figure~\ref{fig:training-gnorm}).

\end{document}